\documentclass{article}

\usepackage{iclr2024_conference}
\iclrfinaltrue
\usepackage{times}
\usepackage[utf8]{inputenc} % allow utf-8 input
\usepackage[T1]{fontenc}    % use 8-bit T1 fonts
\usepackage{hyperref}       % hyperlinks
\usepackage{url}            % simple URL typesetting
\usepackage{booktabs}       % professional-quality tables
\usepackage{amsfonts}       % blackboard math symbols
\usepackage{nicefrac}       % compact symbols for 1/2, etc.
\usepackage{microtype}      % microtypography
\usepackage{xcolor}         % colors
\usepackage{enumitem}
\usepackage{caption, subcaption}
\usepackage{graphicx}
\usepackage{amsmath}
\usepackage{amssymb}
\usepackage{mathtools}
\usepackage{amsthm}
\usepackage{files_iclr/picins}% http://ctan.org/pkg/picins

\DeclareMathAlphabet{\pazocal}{OMS}{zplm}{m}{n}
\theoremstyle{definition}
\newtheorem{definition}{Definition}

\theoremstyle{theorem}
\newtheorem{theorem}{Theorem}

\def\z {\mathbf{z}}
\def\y {\mathbf{y}}
\def\ty {\tilde{\mathbf{y}}}

\def\hy {\hat{\mathbf{y}}}

\def\x {\mathbf{x}}

\def\u {\mathbf{u}}

\def\X {\mathbf{X}}
\def\Y {\mathbf{Y}}

\def\Z {\mathbf{Z}}
\def\U {\mathbf{U}}

\def\th{\mathbf{\theta}}
\def\ph{\mathbf{\phi}}
\def \expec {\mathop{\mathbb{E}}}

\def\Th{\mathbf{\Theta}}
\def\th{ \pmb \theta}

\newcommand{\indep}{\rotatebox[origin=c]{90}{$\models$}}

\def \Zspace { \mathcal{Z} }
\def \Xspace { \mathcal{X} }
\def \Reg {\rho}
\def \L {\pazocal{L} }

\def \Rspace { \mathbb{R}}
\def \Xspace { \mathcal{X} }

\title{Representation Disentanglement via \\ Regularization by Causal Identification }

\author{%
  Juan~Castorena \\
  CCS-3 Information Sciences\\
  Los Alamos National Laboratory\\
  Los Alamos, NM, 87545 \\
  \texttt{jcastorena@lanl.gov} \\
}

\begin{document}

\maketitle

\begin{abstract}
%This work focuses on the problem of learning disentangled representations from observational data. 
%Given dataset observations $\{\x^{(i)}\}_{i=1}^N$ drawn from $p(\x|\y)$ with generative variables $\y$ admitting the distribution factorization $p(\y) = \prod_{c} p(\y_c )$, we ask whether learning disentangled representations is plausible. 
%In this work, we argue modern deep representation learning models for disentanglement are ill-posed with collider bias behaviour; a source of bias producing dependencies between the underlying generating variables.
%Under the rubric of causal inference, we show this issue can be explained and reconciled under the condition of causal identification; attainable from data and a combination of constraints, aimed at controlling the dependencies characteristic of the causal graphical \textit{collider} model encoding the data generation process assumptions.
In this work, we propose the use of a causal collider structured model to describe the underlying data generative process assumptions in disentangled representation learning. %, often assumed to satisfy the i.i.d. factorization property $p(\y) = \prod_i p(\y_i)$. 
This extends the conventional i.i.d. factorization assumption model $p(\y) = \prod_i p(\y_i)$, inadequate to handle learning from biased datasets (e.g., with sampling selection bias).
%The collider structure, explains the plausibility to learn conditional dependencies between the underlying generating variables, even when these are in reality unrelated, creating challenges for disentanglement. %, even when these are in reality unrelated.
The collider structure, explains that conditional dependencies between the underlying generating variables may be exist, even when these are in reality unrelated, complicating disentanglement. 
Under the rubric of causal inference, we show this issue can be reconciled under the condition of causal identification; attainable from data and a combination of constraints, aimed at controlling the dependencies characteristic of the \textit{collider} model. % encoding the underlying data generation process assumptions.
%under supervision or a weak-form of it. 
For this, we propose regularization by identification (ReI), a modular regularization engine designed to align the behavior of large scale generative models with the disentanglement constraints imposed by causal identification. %domain knowledge encoded by a graphical causal model. 
Empirical evidence on standard benchmarks demonstrates the superiority of ReI 
%as applied to a variational framework in removing the effects of collider-bias. 
in learning disentangled representations in a variational framework.
In a real-world dataset we additionally show that our framework, results in interpretable representations robust to out-of-distribution examples and that align with the true expected effect from domain knowledge. %between the generating variables and measurement apparatus. 
%This work focuses on the problem of learning disentangled representations from observational data. Given dataset observations $\{\x^{(i)}\}_{i=1}^N$ drawn from $p(\x|\y)$ with generative variables $\y$ admitting the distribution factorization $p(\y) = \prod_{c} p(\y_c )$, we ask whether learning disentangled representations is plausible.
%Under the rubric of causality, we make the connection that disentanglement of the generating factors $\y_c$ is possible whenever they are $d$-separated. We argue modern disentanglement representation learning models are ill-posed with collider bias behaviour; a source of bias producing entanglement between the generating variables.   
%We show the underlying issue behind representation entanglement, present in modern learning frameworks can be explained through a graphical collider-based data generation model.  which can be reconciled through identification by $d$-separation between variables of the condition of identifiability; attainable under supervision or a weak-form of it. For this, we propose regularization by identification (ReI), a modular regularization engine designed to align large scale generative DL model behavior with identification of $d$-separation conditions made explicit via the causal graphical model. Empirical evidence shows that enforcing ReI in a variational framework results in interpretable disentangled representations equipped with generalization capabilities to out-of-distribution examples and that aligns with the true expected effect from domain knowledge between generating variables and measurement apparatus. 
\end{abstract}

\section{Introduction}\label{Sec:intro} 

One of the principal objectives of learning representations has been that of detecting measurement features that represent the qualitative and quantitative characteristics of the underlying physical processes being sensed. Most of the times sensing as dictated for example by the Nyquist rate \cite{shannon1948mathematical} acquires sufficient information for detection but leaves potentially unnecessary and redundant information on its measurements. Ideas to reduce such redundancies by representing information as concepts, patterns or features to achieve an economy of information have been the focus of study since its early days \cite{pearson1901}. 
%Classical examples include principal component analysis \cite{pearson1901liii} which assumes linear mapping functions along with orthogonality in its parametrizations. Independent component analysis \cite{bell1995information} on the other hand, restricts the elements of the parametrizations to be independent using mutual information. However, both of these methods assume all measurements $\x$ live in a low dimensional space, an assumption not applicable in all task contexts.
%
% Mathematical formulation
Very recently, variational formulations for representation learning such as the variational autoencoder (VAE) \cite{kingma2013auto} and denoising diffusion probabilistic models (DDPM's) \cite{ho2020denoising} have been among the most popular methods. %, very recently.
%The standard variational formulation of  representation learning frameworks consists in learning from the observables $\x \in \Rspace^N$ a generative model $p_{\th}(\x, \z)=p_{\th}(\x | \z) p_{\th}(\z)$ whose learned marginal likelihood $p_{\th}(\x)$ approximates the true $p_{\th^*}(\x)$. The latent variables $\z \in \Rspace^d$ distributed as $p_{}(\z)$ are assumed unknown and focus on this problem has concentrated on finding priors parametrized by $\ph$ that make the marginal and posterior $p_{\ph}(\z | \x)$ tractable. 
% VAE's
%Variational auto-encoders (VAE)'s \cite{kingma2013auto} for example, builds an efficient optimization approach that maximizes the conditional likelihood $p_{\th}(\x|\z)$ subject to similarity constraints quantified by the Kullback-Leibler (KL) divergence between a posterior approximate $q_{\ph}(\z|\x)$ and a family of induced latent distribution priors $p(\z)$ (e.g., an isotropic Gaussian). 
% Problem
Problem with these group of methods %for disentanglement 
is their focus on learning approximations of the true marginal data distributions without any guarantees on the imposed representation priors to model the true underlying generative mechanisms \cite{khemakhem2020variational}. 
This not only disconnects the learned latent representations from real-meaning, obfuscating explainability of the generative process \cite{lake2017building} and counterfactual reasoning \cite{pearl2019seven}, but also invokes problems of fairness and robustness to out-of-distribution (OOD) examples \cite{d2020underspecification}.
Recent trends \cite{bengio2013representation, higgins2018towards,  locatello2019challenging, van2019disentangled, khemakhem2020variational} are in consensus that disentanglement of the generating factors leads to increased robustness, explainability and fairness.
%robust representations that are less susceptible to the appearance of subsets of entangled factor variables in the intended tasks.
%; with ideas rooted in causality \cite{pearl1994probabilistic, pearl1995causal}.
% Definition of disentanglement 
%The most common notion of disentanglement assumes a set of generative latent factors that explains the data has a one-to-one correspondence between each factor and a single (or a subset of) dimension(s) of the learned representations \cite{bengio2013representation, higgins2016beta, chen2018isolating, eastwood2018framework, trauble2021disentangled, roth2022disentanglement}.
The most widely used definition of disentangled representations assumes the set of underlying generative factors that explains the data has a one-to-one correspondence between each factor and a single (or a subset of) dimension(s) of the learned latent representations \cite{bengio2013representation, higgins2016beta, chen2018isolating, eastwood2018framework, trauble2021disentangled, roth2022disentanglement}.
%In this sense, the properties of disentanglement are established as characteristics of the encoding representation heuristics. %This is a relaxed definition of each dimension capturing only one factor of variation as in \cite{suter2019robustly}. 
%
%Recent efforts along this line of work, includes unsupervised learning methods that exploit the enormous amounts of data available without the requirement of labels for each generating factor. 
% Unsupervised Methods
Recent efforts along this line of work, include unsupervised methods that rely on encoder heuristics to control the information bottleneck properties for disentanglement. %For example, by encoding inputs into a lower dimensional space.
%Methods relying on the information bottleneck assumption rely on a heuristic on the regularization strength $\lambda$ restricting the encoder characteristics that establish a tradeoff between an information bottleneck and the input distribution likelihood. 
%Careful tunning of encoder hyper parameters needs to be considered for disentanglement to be effective for each specific distribution of observations and true generating factors. Here instead, our 
Among the most popular VAE based methods includes $\beta$-VAE \cite{higgins2016beta}, Annealed VAE \cite{burgess2018understanding}, Factor VAE \cite{kim2018disentangling}, DIP-VAE \cite{kumar2018variational} all imposing specific structure in the latent prior through modifications of the Kullback-Leibler divergence (KL) term. %The $\beta$-VAE \cite{higgins2016beta} for example includes a $\beta$ scalar that controls the strength enforcing the latent prior. 
%Increasing this scalar promotes the structure of the prior at the cost of divergence from the true marginal likelihood, thus $\beta$ being a balancing term. 
On the denoising diffusion front, the work of \cite{yang2023disdiff} instead minimizes the mutual information between latent representations of an autoencoder and uses it in a guided denoising diffusion framework.
These unsupervised methods, rely on careful tunning of the encoder hyper-parameters to preserve the desirable features in the data while destroying features of nuisance factors or of no particular interest.
%One drawback, is that careful tunning of encoder hyper-parameters needs to be considered for disentanglement to be effective for each dataset. %for each specific data distribution and underlying generating factors. 
%Moreover, rooted in the concept of model identifiability, \cite{locatello2019challenging} challenged the line of work of unsupervised learning disentanglements as impossible.
% Weakly supervised learning
Weakly-supervised methods on the other hand, %have exploited weak-labels as they 
have been shown to facilitate some form of disentanglement.
%Other lines of work operating through a more principled approach of analysis under the lens of causality rely on achieving some notion of model identifiability of the generating mechanisms, at least implicitly. Identification provides the guarantees that a given  {\color{red} Why is identification important?.}
 \cite{mitrovic2020representation} proposes a method that consists in learning representations explaining the causal data generation mechanisms by promoting invariance to augmented data transformations, this under the principle of independent causal mechanism \cite{peters2016causal, peters2017elements}. The goal of invariant risk minimization (IRM) of \cite{arjovsky2019invariant, rojas2018invariant}, on the other hand, is to find representations that produce predictions invariant to environment contexts. The work of \cite{lu2021invariant} further extends this to the non-linear learning setting. \cite{bouchacourt2018multi} proposes learning representations by grouped observations (i.e., a factor of variation shared between observations within a group) and uses a multi-level VAE for learning group representations as a generalization to i.i.d. assumptions. % and tackle the limitations of VAE's assuming i.i.d observations.
%
%Mencionamos aqui tambien el trabajo de \cite{mao2022causal} el cual su enfoque no es el de aprendizaje de representaciones si no el de construir bajo causalidad  neural machines with OOD transportability capabilities. La idea es similar a las ya mencionadas en las cuales se leverage la variabilidad en los spurious factors cuando estas provienen de labels en comun.
%
\cite{locatello2020weakly} demonstrates that disentangled representations can be obtained under weak-supervision when pairs of measurements share a factor of variation. Their approach modifies the $\beta$-VAE objetive by enforcing similarities between the shared generative factors of variation and a decoupling of those uncommon. Worth noting is \cite{trauble2021disentangled, roth2022disentanglement}, whose findings extend disentanglement to cases with correlated factors; a problem that affects robustness to OOD examples \cite{d2020underspecification}. 
In light of these works, the scope of this research is to learn representations that disentangle the underlying generative factors of variation.
Here, we follow the line of work of \cite{suter2019robustly, locatello2020weakly, lu2021invariant} where the framework for disentanglement is connected to the underlying causal mechanisms explaining the data generation process.
%The line of work consists on imposing causal-effect identification \cite{pearl1995causal} constraints based on the data generative mechanism model. Different from the works in \cite{peters2017elements, bouchacourt2018multi, arjovsky2019invariant, locatello2020weakly, mitrovic2020representation, khemakhem2020variational}, this approach adopts a Bayesian's view of the model parameters $\th$ and argues that for representation disentanglement in distributed large scale models \cite{hinton1984distributed, hinton1990connectionist} not only the latent variables need to be controlled for identification, but also the model parameters. We argue that distributed learning over many neural units without control, a paradigm used in many DL approaches, generates difficulties for disentanglement of the interplaying generative factor variables. Learning updates throughout the entire deep model without control allows the free-paths of information flow through it. Our framework aims at blocking causally unsupported paths to help the DL model avoid making associations between causally unsuported factor relationships. %All this without getting into the century old dilemma between frequentist and Bayesians \cite{greenland2006bayesian} on interpretations of the mechanistic nature reflected by the model parameters, but rather on reflecting on the nature of the uncertainty about $\th$. 
Our main contributions are:
\begin{itemize} [noitemsep, leftmargin=*]
    \item Provide a connection between the definition of disentanglement and causal identification constraints which are informed by graphical causal models that encode the underlying data generation process.
    %: imposing disentanglement constraints that control dependencies between the underlying generating factors encoded by a graphical model.
	\item %The generative model admitting the distribution factorization $p(\y) = \prod_{c} p(\y_c )$ as used by many popular disentanglement approaches \cite{higgins2016beta, higgins2018towards, kim2018disentangling, chen2018isolating} and its relaxed statistically dependent factor counterpart \cite{trauble2021disentangled, suter2019robustly,  roth2022disentanglement}, we argue, exhibits collider bias behavior. 
    %We argue modern approaches for learning disentangled representations exhibit collider-bias behavior. 
    We propose the use of a causal collider structured model to describe the underlying data generative process asumptions in disentangled representation learning. 
    %This is a type of bias, characterized under graphical causal models by a collider structure, that explains the appearance of conditional associations between the generating factors, when observing the common effect, even when factors are in reality unrelated, thus producing entanglement.  
    %This is a type of bias, characterized under graphical causal models by a collider structure, that explains the appearance of conditional associations between the generating factors, even when these are in reality unrelated, thus producing entanglement. 
    This collider model, can explain entanglement in the learned representations by the appearance of conditional dependencies between the generating factors, even when these are in reality unrelated.
    %This is novel to the best of our knowledge. 
    %Moreover, contrary to the general exposition that deep learning frameworks have a propensity to exploit shortcuts \cite{beery2018recognition, geirhos2020shortcut, pezeshki2021gradient}, we rather propose explaining these effects as collider bias behaviour.
	%We show under a pre-specified DAG model $G$ encoding the typical data generative process used in the disentanglement works of \cite{higgins2016beta, higgins2018towards, kim2018disentangling}, that standard representation learning frameworks present collider bias behaviour; a type of bias producing entanglement between the effects of the generating factors. 
%\item Demonstrate the inductive bias in standard distributed deep model architectures contain mechanisms that facilitate the free flow of information, destroying or preventing control over the generative variables as required by identification, thus resulting in entaglement.  

	\item ``Regularization by Identification" (ReI), a modular regularization engine designed to align the behavior of large scale DL models with causal identification constraints. This, enforces disentanglement by controlling dependencies between the underlying generating factors.
	\item A variational inference reformulation of the VAE representation learning problem (i.e., the ELBO) to achieve disentanglement by imposing ReI under the collider graphical model.
    %that guarantee identification and that ultimately produce disentaglement. {\color{red} Somewhere in the text include that we corroborate the findings of \cite{locatello2019challenging} that disentanglement is directly connected to identification.}
	\item Provide empirical evidence from both disentanglement benchmarks and real-world datasets showing the potential of ReI %in a dataset with joint variability between the generating factors; an issue recognized by \cite{trauble2021disentangled, roth2022disentanglement} as the problem of correlated data. In this case, ReI offers the potential 
    to produce representations that: (1) disentangle the effects of the generating factors with results well aligned with true expected behavior from domain knowledge that support interpretation and understanding and (2) are robust in the presence of out-of-distribution examples. %in comparison to the standard non-identifiable DL model counterparts.
	%Applications of the proposed causally driven regularizations, results in representations that align with the characteristic spectral response of each chemical element. In addition, the exerted causal regularization control results in more explainable and robust representation in the presence of out-of-distribution examples compared to the uncontrained deep learning methods even when our causally regularized neural-net is singled layer.
\end{itemize}

\vspace{0.5em}							% Introduction

\section{Learning Disentangled Representations}
%
%\begin{enumerate}[nolistsep]
%	\item Generative factors $p(\y) = \prod_{i} p(\y_i )$
%	\item Marginal data distribution $p(\x)$ and joint data distribution $p(\x,\y)$
%	\item Variational inference framework graphical generative model. Problem when conditioning either on $\z$ or on $\x$.
%\end{enumerate}
\textit{Generative Representation Learning:} Consider observations $\x \in \Xspace  \subseteq \Rspace^M$ drawn from distribution $\sim p^*(\x)$. The goal of generative models for representation learning is to find encoders $q_{\th}: \Xspace \rightarrow \Zspace$ that produce latent representations $\z \in \Zspace \subseteq \Rspace^d$ distributed according to some prior distribution $p(\z)$, that along with a generator $p_{\ph}: \Zspace \rightarrow \Xspace$ marginally approximates the input data distribution.

%\textit{Disentanglement:} The most common notion of disentanglement assumes a set of generative latent factors that explains the data has a one-to-one correspondence between each factor and a single (or a subset of) dimension(s) of the learned representations \cite{bengio2013representation, higgins2016beta, chen2018isolating, eastwood2018framework, trauble2021disentangled, roth2022disentanglement}. 
%{\color{red} TODO: Include a small discussion about disentanglement including some examples.}

%\subsection{Connection between Disentanglement and Causal Effect Identification}

%In contrast to standard approaches, we consider a causal inference model based definition of disentanglement where domain knowledge about the data generation process is exploited, following the works of \cite{suter2019robustly,shen2022weakly}. 

\subsection{Causal Inference Background}
\begin{definition}
\textit{Directed Acyclic Graphs:}
%
% What is a DAG?
In causal inference, the data generation process is represented by a directed acyclic graph (DAG) \cite{pearl1995causal}. A DAG is a graphical model with domain variables represented as nodes, directed edges (i.e., arrows) expressing directional dependency relationships between variables.
%
% Markov compatibility: DAG's integrate statistical information
DAG's operate under the Markov compatibility property which states that the joint distribution $p$ is compatible with a DAG $G$ or that $G$ represents $p$ if it admits the decomposition
\begin{equation} \label{markov}
p(x_1, ..., x_n) = \prod_i p(x_i | pa_i).
\end{equation}
Variables $pa_i$ are the Markovian parents of node $x_i$ that belong to the minimal set of predecessors that renders $x_i$ independent of all its other predecessors; in other words  that, $p(x_i|pa_i) = p(x_i| x_1, ..., x_{i-1} )$ \cite{Pearl:2010}. 
Parents $pa$ and predecessors are defined along the arrows in the graph. For example, in $X \rightarrow Z \rightarrow Y $, $X$ is the only parent of $Z$, $Z$ is the parent of $Y$ and $\{X,Z\}$ is the list of predecessors of $Y$.
We note that this Markov compatibility assumes a first order Markov process as defined in Eq.\eqref{markov}.
% Benefits of using such graph models
%This Markov property restricts the unbounded number of plausible models that can fit the joint distribution to only those compatible with the specified DAG. 
%decomposition. 
% Expand a little more on dependencies
%Some variables affect outcomes, some affect observed inputs (or treatments), and some affect both. There can be three basic types of dependencies in DAG's: (1) a mediating node between two nodes, (2) a common cause which shows a mutual dependency between two nodes on a third and (3) a common effect on a node mutually affected by two other nodes \cite{}. 
Causal paths between input $X$ and outcome $Y$ consist of a sequence of arrows following the causal direction and represents causal dependencies. 
Non-causal paths, on the other hand, consist of a sequence of connections between variables that lack a direct cause-and-effect relationship and represent dependencies that may arise from non-causal influences. 
The takeaway is that non-causal dependencies without control produce biased estimates.
Examples include confounding from shared causes or collider-bias arising by conditioning on a common effect.
\end{definition}

% Causal inference Analysis
Causal inference analysis provides the tools %to predict causal effects 
%causal inference analysis provides the axioms of $do$-calculus (included in Appendix \ref{Ssec:axioms}) 
to establish causality by predicting the effects of interventions $p(y|do(x))$ from the assumed DAG $G$ and ordinary distributions over observations. The way by which this is accomplished is by removing the effects of any non-causal dependencies %(i.e., the effect of variables in non-causal paths) 
between the input and output.
%However, under causal interpretation, DAG's are not only about this Markov restriction shared also by Bayesian networks \cite{pearl1988probabilistic}, but more than that, they aim to move relationships between variables beyond mere correlation into causation. The way by which this is accomplished is by removing the effects of any non-causal dependencies (i.e., the effect of variables in non-causal paths) between variables.
%
% What is the DAG used for?
%At the core of ReI is the use of graphical models (i.e., DAG's) \cite{pearl1995causal}, which provide the mathematical language for expressing domain knowledge through \textit{transparent} and \textit{testable} assumptions about the underlying causal relationships between variables. 
%The DAG specifies all the assumptions that we have made about the problem. For example, what variables from domain knowledge are we considering as fundamental, or which variables are observable (measured) and which are not (i.e. latent). 
%Also, how variables interact with each other (i.e., dependence, independence, conditional independence) and what is the assumed direction of causality within those variables. 
%This of course can be at different levels of abstraction, but the key here is that the graphical model is not only a simplification but rather a specification of all the assumptions made about the underlying physical data generation process.
The DAG provides the mathematical language for expressing domain knowledge through transparent and testable assumptions about the underlying relationships between domain variables. Transparency enables analysts to discern whether the stated assumptions 
%encoded qualitatively in the DAG $G$ 
are plausible \cite{pearl2019seven} (on scientific grounds). Testability provides graphical criteria to determine the causal/non-causal dependencies between variables. If the non-causal influences can be controlled, then it provides the rules to do so. % and identify the causal effect.
At the core of these causal tools, is the $d$-separation criteria \cite{geiger1990d}. %to test for dependencies between variables. 
It is graphical-based, in the sense that the structure of the DAG $G$ (i.e., edge connections and paths) encodes qualitatively the patterns of dependencies we should expect to find in the data. %These dependencies may exist either from influence of connected variables along the causal paths or from an entire system of interconnected non-causal paths. 
%The graphical criteria of $d$-separation tests the structure of 
In addition, it provides the means to control for any dependencies through conditioning by an appropriate set of covariates $\Z$.
\begin{definition} \label{def:d-separation}
Variable sets $\X$ and $\Y$ are $d$-Separated (or blocked) by a set $\Z$ denoted as $(\X \; \indep \; \Y | \Z)_G$, if and only if, $\Z$ blocks all paths from nodes in $\X$ to nodes in $\Y$ \cite{geiger1990d,pearl1995causal}. The two general graphical conditions for blocking dependencies are:
\begin{itemize} [topsep=0pt,itemsep=-1ex,partopsep=1ex,parsep=1ex, leftmargin=*]
	\item  In the paths $\X \rightarrow m \rightarrow \Y$ or $\X \leftarrow m \rightarrow \Y$ the node $m$ is in $\Z$, or
	\item there is a collider $\X \rightarrow m \leftarrow \Y$ where neither node $m$ nor its descendant is in $\Z$.
\end{itemize}
When no feasible set $\Z$ exists then we say that $\X$ and $\Y$ are not $d$-separated.  When it does then we can control for dependencies by conditioning on $\Z$.
%and dependencies exist between these variables in almost all distributions compatible with the DAG.
Definition \ref{def:d-separation} explicates that the $d$-separation criteria provides the graphical test to determine the dependencies that exist in the system of variables outside of input $\X$ and outcome $\Y$, and the mechanisms to control for these dependencies.
\end{definition} 
\begin{theorem} \label{probabilistic-d-separations}
    \cite{verma1990causal} \textit{Probabilistic implications of $d$-Separation.} $(\X \; \indep \; \Y | \Z)_G$ implies conditional independence of $\X$ and $\Y$ given a set of variables $\Z$ (including $\Z = \{\emptyset\}$) in every distribution compatible with the encoded assumptions in DAG $G$, while absence of $d$-Separation implies the converse; a dependence in almost all distributions compatible with the DAG. %\cite{geiger1990d}. 
\end{theorem}
%
%Las implicaciones del teorema 1, nos permiten controlar las dependencias que existan entre dos variables si estas estan $d$-separadas. Asi mismo, estas dependencias estan garantizadas de permanecer controladas bajo todas las distribuciones compatibles con la DAG especificada. En Pearl \cite{pearl1995causal} se establece que esto, nos permite determinar los efectos causales entre dos variables bajo la condicion de identificacion.
The implications of Theorem \ref{probabilistic-d-separations} allow us to control the dependencies that exist between two variables if they can be $d$-separated. Likewise, these dependencies are guaranteed to remain controlled under all distributions compatible with the specified DAG. In \cite{pearl1995causal} it is established that this, allows us to identify the causal effects between two variables under a specified DAG $G$ and data.

%
%Consecuentemente, bajo la independencia condicional de $d$-separation en todas las distribuciones compatibles con el DAG permite que   
%
%In addition to testing, $d$-separation provides a link between the selection of an appropriate set of covariates $\Z$ required to block or shield for non-causal dependencies. Variables $X$ and $Y$ are $d$-Separated given set $\Z$ if and only if $\Z$ blocks every path (i.e., a set of consecutive connected edges) from node $X$ to node $Y$ \cite{pearl1995causal}. 

%Based on $d$-separation, causal inference analysis provides the axioms of $do$-calculus (included in Appendix \ref{Ssec:axioms}) to establish causality by predicting the effects of interventions $p(y|do(x))$ from the assumed DAG $G$ and ordinary distributions of observations. 
%The effects of interventions written as $do(x=x)$ represent how external agents may affect the system by forcing certain variables to take on predefined values. 
%
\begin{definition} \label{def:identification}
    \textit{Causal effect Identification \cite{pearl1995causal}}. The causal effect of $\X$ on $\Y$ denoted as $p(\Y|do(\X))$ is identifiable from DAG $G$ and data, if a set of variables $\Z$ that $d$-separates them exists.
\end{definition}
The implications of identification state that dependencies found in the DAG can be controlled through a combination of $d$-separation constraints. If the causal effect is identifiable, then this control is guaranteed to hold in every distribution compatible with the DAG, while also licensing computations from the joint distribution $p(\X,\Y,\Z)$ over the observables. In other words, the causal effect in the l.h.s. of $p(\Y|do(\X=\x)) = \sum_{\Z} p(\Y|\x,\Z) p(\Z)$) can be computed through the r.h.s. involving only standard distributions over the observations. \cite{ayemreview} overviews methods for causal identification.

\subsection{Connection between Disentanglement and causal identification}

\textit{Learning Disentangled Representations by Causal Identification}. Given a DAG $G$ encoding our assumptions about the underlying data generative process, the possibility to learn disentangled representations from data, we propose, 
%with respect to the underlying generating factors from the joint distribution over the observations 
can be tested through $d$-separation of the generating factors. Moreover, identification of the causal effect $p(\x | do(\y_c))$ for all factors $\y_c, c \in \{1,...,n\}$ provides the necessary conditions to control for dependencies between the generating factors, through a combination of $d$-separation constraints. % and data drawn from the joint distribution over the observables.
%
%\textit{$d$-separation between Generative factors}. A data generative process encoded by a given DAG $G$ is $d$-separated with respect to pairs of the generative factors $\y_c$ and $\y_j$, for all $c \neq j$, ,if and only if, the generative factors are $d$-separated given an appropriate set $\Z$. %In other words, $\y_c ~ \indep ~ \y_j | \Z$.
% Implications of d-separation on generative factor distributions.
%La primera proposicion que hacemos es la de el uso de el criterio de $d$-separation en los procesos de generacion de datos encoded as a DAG para $d$-separar las variables generativas y remover las dependencias que existan entre ellas.
% 
Given a dataset whose generative process is transparently encoded as a DAG with nodes representing the generative variables and edges the assumed relationships between them, the proposition involves to first qualitatively \textit{test} whether the generative variables can be $d$-separated relative to each other. 
From Theorem 1, the possibility to identify the causal effect of each generative variable through $d$-separation implies that the dependencies between the generative variables can be controlled in every distribution compatible with the specified DAG.
This, we propose, enables disentanglement of the generative factors from a DAG $G$ and data with guarantees that hold in every distribution that satisfies the DAG compatibility. 
%{\color{red} TODO: Connection between this and the definition of disentanglement.} 
The converse: absence of $d$-separation, implies dependencies between the generating factors in almost all distributions compatible with the DAG.
%In addition, from Theorem \ref{probabilistic-d-separations}, in the absence of $d$-separation there will be a dependence between the generating factors in almost all distributions compatible with the DAG. 
The later, potentially leaves learning by DL models free to exploit any correlations available in the data, which can be problematic for disentanglement, specially considering their susceptibility to exploit shortcuts \cite{beery2018recognition, geirhos2020shortcut, pezeshki2021gradient}.
%Without control of these dependencies, leaves DL models free to exploit any of such dependencies, which can be problematic for disentanglement, specially considering the susceptibility of DL models to exploit shortcuts \cite{beery2018recognition, geirhos2020shortcut, pezeshki2021gradient}. 
%Our proposition here instead, provides first a graphical qualitative test to describe the dependencies between variables expected  in the data generating process and second, the conditions required to control dependencies of the generating factors which are guaranteed to hold under all distributions compatible with the DAG. 
Our proposition here instead, %provides first a graphical qualitative test to describe the dependencies between variables expected  in the data generating process and second, the conditions required to control them with guarantees that hold under all distributions compatible with the DAG.
%This later, 
imposes control over the shortcuts explored by the DL model by constraining dependencies between the underlying generating factors. 
%By definition, identification of the causal effects involves $d$-separation between the generative factors $\y_c ~ \indep ~ \y_j | \Z$ implies conditional independence of $\y_c$ and $\y_j$ given $\Z$ in every distribution compatible with the DAG $G$. Conversely, the absence of $d$-separation between generating variables implies dependencies in almost all distributions compatible with $G$. We propose the use of $d$-separation as the non-parametric criteria to test and ensure disentanglement of the underlying generating variables as follows:
%The proposition implies that disentanglement amounts to a graphical test, given DAG $G$, to check whether the causal effect is identifiable. The answer can be affirmative, in which case, disentanglement is possible and the mechanisms to control for non-causal dependencies in the system can be obtained through $p(\x | do(\y_c))$. On the other hand, if no set $\Z_c$ exists, then $p(\x | do(\y_c))$ is not identifiable from the available observations alone and more experiments or measurements are required.

% Difference between this definition of disentanglement and other standard notions of disentanglement.
%Point out the differences between this causal-based definition and standard notions of disentanglement \cite{} is ...... Methods operating under causal principles \cite{suter2019robustly, shen2022weakly} differ by ....

\subsection{Directed Acyclic Graph: Collider-based Data Generation Model}
At the core of causal identification, is the reliance on a graphical DAG model encoding the data generation process to identify (if possible) causal effects, and consequently remove dependencies between the generating factors producing entanglement.
Here, we propose a simple DAG model that yet, explains and reconciles the entanglement between the generative factor effects on data.

\pichskip{15pt}% Horizontal gap between picture and text
\parpic[r][t]{%
	\begin{minipage}{35mm}
		\includegraphics[width=\linewidth]{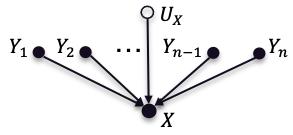}%
		\captionof{figure}{{\small{DAG $G$ encoding data generative process.}}}
		\label{fig:sensor_DAG}
	\end{minipage}
}
%
%\begin{proposition}
    \textit{Collider Data Generative Model.} 
    We argue that the underlying data generative model for disentanglement has collider structure. Generating variables (i.e., causes) $\y_c, \forall c \in \{1,...,n\}$ have directed arrows colliding at node $\x$ (i.e., common effect). The DAG representing this simple, yet generic structural model is illustrated in Fig.\ref{fig:sensor_DAG}
%\end{proposition}
%
%\vspace{-1em}
%

Variable realization $\x \in \Rspace^M$ represents the sensor measurements (e.g., image), $\y \in \Rspace^n$ with elements $\y_c, c \in \{1,...,n\}$ are the underlying generating factors (permutations are possible as long as structure relationships are preserved). Nuisance factors (e.g., sensor noise, style), which are not relevant to the tasks we care about, are denoted by the unmeasured $\u_x \in \Rspace^M$. 
Arrows emanating from $\y_c, ~\forall c$ to $\x$ (i.e., colliding at $\x$) are aligned with the causality of the generation mechanisms. In other words, there is causal precedence of the $\y_c$'s and they are a direct cause (i.e., implied by the arrow direction) of the common effect $\x$. 
%All arrows collide at $\x$, hence its name \cite{berkson1946limitations, kim1983computational, pearl2009causality}. 
%This indicates that the parent nodes $\y_c$ are the generators of $\z$ and that these are the only dependencies of collider $\z$. 
%The bi-directed arcs $\y_i \leftrightarrow \y_j$ for $i \neq j$ represent dependent unmeasured effects between the edge connected variables, making this model semi-Markovian. %One case that presents such dependencies for example is when the connected variables grouped together for sensor measurement become dependent. 

Inspection of Fig. \ref{fig:sensor_DAG} reveals potential dependencies arising from the structural connections $\y_c \rightarrow \x \leftarrow \y_j$ with $j \neq c $, indicative of a collider; a source of potential bias \cite{berkson1946limitations, kim1983computational, pearl2009causality}. This bias occurs when two or more independent variables have a direct causal influence on a variable as they will become associated when conditioning on (e.g., observing) the common effect $\x$. This spurious conditional association between the generating factors is a source of entanglement if it remains without control.
% Intuition
For example, consider the case of $C = A + B$ with $A$ and $B$ independent. When variable $C$ is observed (e.g., $C=4$) variables $A$ and $B$ become conditionally correlated as knowledge of one informs the value of the second. 
Intuitively, this phenomenon occurs as information on one of the causes makes the other causes involved more or less likely given that the consequence has occurred (i.e.,  the explaining away
effect \cite{pearl1988embracing}); even when the causes are independent \cite{pearl1995causal}. 
Such behavior has been observed in the context of data-based models also, but has been rather explained as a susceptibility to exploit shortcuts \cite{beery2018recognition, geirhos2020shortcut, pezeshki2021gradient}.  
%In this case, For example, a DL model in a visual classification task can learn conditional associations between class-specific features and background, given an observed image, failing to generalize when such associations do not hold. Specifically, learning classifiers for cats and lions from images could associate lions with backgrounds in the wild  and cats with backgrounds in domestic settings. Similar problems have been reported in numerous works \cite{}. 
Here, we argue such behavior can be explained instead as collider-bias from an underlying collider structured data generative process. 

% It would be useful to demonstrate with benchmark datasets (e.g., shapes3d) that leaving even one factor without control, results in entanglement between the generating factors.

% Controlling collider variables
Controlling for conditional dependencies between a single $\y_c$ and the remaining generating factors, requires finding the set that $d$-separates them. %supervision or a weak form of it. 
%Otherwise, if paths remain open will result in entangled representations. 
Inspection of Fig.\ref{fig:sensor_DAG} reveals that $(\y_c ~ \indep ~ \x | \Z)_G$ with $\Z = \{ \y_j : j \neq c\} \cup \{\u_{\x} \}$. Causal identification $p(\x|do(\y_c))$, $\forall c$, %controlling for entanglement between the generating variables 
is thus given as:
\begin{equation} \label{collider-control}
p(\x | do(\y_c))
 =  \sum_{\y_{-c},\u_x} p(\x | \y, \u_x) p(\y_{-c},\u_x) = \expec_{p(\y_{-c},\u_x)} \left [ p(\x|\y, \u_x)\right]
 \end{equation}
In other words, conditioning over $\y_{-c}= \{ \y_j : j \neq c\}$ which denotes the generative factors in the system except $\y_c$. %, with $\y = \y_{-c} \cup \{\y_c\}$.
Full derivation of the expression in Eq.\eqref{collider-control} can be found in Appendix \ref{appendix:identification}. Note that controlling for dependencies between the generating variables requires conditioning, which in turn requires some form of measurements of both $\y$ and $\u_{\x}$. 
Identification is possible thus, given supervisory signals, or a weak-form of it \cite{shu2019weakly} for both $\y$ and $\u_{\x}$. 
%Conversely, the impossibility to identify causal effects occurs when no measurements from $\y$ and $\u_{\x}$ are available. 
In cases when no measurements or proxies are available, dependencies between the involved variables that are compatible with the specified DAG will be present and representations will in turn present entanglements between the effects of such dependencies. 
For example, if variable $\u_{\x}$ is unmeasured, opens any of the paths $\y_c \rightarrow \x \leftarrow \u_{\x}, ~ \forall c \in \{1,...,n\}$ leaving any plausible association between $\y_c$ and $\u_{\x}$ without control. Similarly, the paths $\y_c \rightarrow \x \leftarrow \u_{\x} \rightarrow \x \leftarrow \y_{-c}$ without control, introduce entanglements between $\y_c, \u_{\x}$ and $\y_{-c}$. 
%TODO: Maybe adding a few toy examples would be nice.
Such collider model thus explains that DL models are free to exploit any of the unbounded number of models compatible with the specified DAG \cite{jaber2018causal} unless some form of control through supervision or a weak form of it is available. 
%We attribute the problems of DL model robustness and fairness to such flexibility to explore such large space of compatible models without explicit constraints to  the underlying data generation process.
%{\color{red} \TODO: Interpretation of Eq.\eqref{collider-control}}. 
%TODO: Also derivation for $p(\x|do(\y))$ and its implications.

% Differences between this model and others in the literature.
In addition to explaining entanglement effects and the means to control them, the collider DAG model is compatible with the independent factors model whose joint distribution admits factorizations of the form $p(\y) = \prod_{c} p(\y_c)$ as in \cite{bouchacourt2018multi, locatello2020weakly, khemakhem2020variational} and also of its correlated factors relaxation, as in \cite{trauble2021disentangled}. 
%In other words, the joint distribution from the DAG in Fig.\ref{fig:sensor_DAG} can be written as $p(\x,\y,\u_{\x}) = p(\x|\y,\u_{\x}) \prod_i p(\y_i) p(\u_{\x})$.
In causal-based disentanglement approaches, graphical DAG models have mostly focused on confounding structures characterized by pairs of generating variables that have a common cause \cite{suter2019robustly, lu2021invariant}. The behaviour of such structures is fundamentally different from those of the collider structure. The former, does not deal with dependencies between the underlying generating factors from conditioning on the common effect. Thus, they are susceptible to this problem, notably in cases when datasets 
%are generated with some factors being more likely to occur than others, 
are not generated by i.i.d. factors, 
as in real-world scenarios. As a remark, \cite{zhang2021adversarial} has explored this collider problem in the context of adversarial robustness.
%{\color{red} TODO: Perhaps describe a little more about the advantages of our model versus the others listed.} 
%Specifically, the causal graphical model we propose to represent the generative model in either of these two cases is structured like a collider and is known by the same name within the framework of causality \cite{kim1983computational}.

%For example, observation of the prescence of a specific object class in a delimited space conditions the prescense of another in that same space. 
%We provide model flexibility to incorporate in a structural graphical way that the generating variables can be associated/correlated in one way or another and with causing factors that may remain unobserved. 
% Problems with collider-like structures

% Joint distribution
%The joint distribution $p(\x,\y, \u_{\x})$ compatible with the collider DAG $G$ in Fig. \ref{fig:sensor_DAG} admits the product decomposition $p(\x|\y,\u_{\x}) \prod_{c} p(\y_c) p(\u_{\x})$

% Variational framework
%In the variational framework, training by enforcing the likelihood term $p(\x|\z)p(\z|\y_c)$ will, we argue open the free flow of information between the colliding generating factors $\y_c$ producing entangled representations if it remains untreated. 

\subsection{Regularization by Causal Identification (ReI)}

%Learning representation models by fitting the distribution over the observables only, yields an unbounded number of plausible models. We could only hope that the fitting function returns a model that aligns with a data generative model with domain knowledge meaning. However, this is rarely the case in practice, even in the presence of just a few generating variables.
%For example, the VAE framework known to be very effective in approximating the marginal data distribution $p(\x)$ does not provide any guarantees that the posterior distribution aligns with the true generating variables, rendering latent representations meaningless in the context of problem domain knowledge. Alternatively, the breadth of recent work most successful in representation disentanglement of the generative factors has relied on an information bottleneck assumption \cite{peters2016causal,bouchacourt2018multi, mitrovic2020representation, locatello2020weakly, khemakhem2020variational, trauble2021disentangled}. However, the information bottleneck provides no guarantees for disentanglement of the generating factors.  

% Bringing back representation learning frameworks.
% DL models learn through a data fidelity objective implied by the likelihood function. 
Representation learning models operate with the objective of fitting the joint distribution over the observations while simultaneously imposing a prior with desired structural characteristics.
The Markov property of DAG's further constrains the unbounded number of plausible models that can fit the joint distribution to only those that are compatible with the specified DAG.
ReI further restricts such models to those that control for entanglement through a reformulation of the learning problem. 
Such reformulation, controls for dependencies between the generating factors by imposing causal effect identification of an assumed collider DAG model as that in Fig.\ref{fig:sensor_DAG}
%Such reformulation, imposes causal effect identification constraints of the generating factors to control for dependencies graphically encoded by a DAG $G$ as in Eq.\eqref{collider-control}. %This process is illustrated in Fig.\ref{fig:data_plus_scm}

%
%\pichskip{15pt}% Horizontal gap between picture and text
%\parpic[r][t]{%
%	\begin{minipage}{55mm}
%		\includegraphics[width=\linewidth]{figures_iclr/data_plus_rei}%
%		\captionof{figure}{{\small{Our approach constraints the unbounded number of DL models, including those that fit a given distribution over the observables, to align with disentanglement by causal identification (i.e., data + ReI).}}}
%		\label{fig:data_plus_scm}
%	\end{minipage}
%}
%Through ReI, we seek to align representation learning frameworks, by a regularization term in the learning problem, imposing disentanglement constraints through causal identification. 

%The weakly-supervised work of \cite{locatello2020weakly} also based on a form of the $\beta$-VAE with shared latent factors between pairs of observations does not recognize this collider source of bias neither and is thus also prone to this problem. {\color{red}Their experimentation however, does not reflect it as datasets were generated by independent factors with no mutual joint variations.}
% and similarly the prediction set $\tW_i = \{ \ty_j : j \neq i\}$, $\tW = \tW_i \cup \{\ty_i\}$. 

%
\textit{Regularization by Identification (ReI)}: is a modular regularization engine designed to align the behavior of DL models by imposing generative factor disentanglement constraints through causal identification.
% Reformulating the learning problem
%Typical representation learning problems consist in learning the posterior $p(\z|\x) = p(\x|\z)p(\z) / p(\x)$ imposing a Gaussian prior $p(\z)$ with independent element constraints. 
%More recent methods for disentanglement rely on encoder heuristics that aim at controlling the information bottleneck properties for disentanglement. For example, by encoding inputs into a lower dimensional space.
%Methods relying on the information bottleneck assumption rely on a heuristic on the regularization strength $\lambda$ restricting the encoder characteristics that establish a tradeoff between an information bottleneck and the input distribution likelihood. 
%Careful tunning of encoder hyper parameters needs to be considered for disentanglement to be effective for each specific distribution of observations and true generating factors. Here instead, our 
%ReI reformulates the representation learning problem to impose disentanglement constraints between the generative factors through causal identification. Given a DAG $G$ of the data generative process, the learning problem uses the causal relationship of $p(\x|do(\y))$ instead of the statistical relationship $p(\x|\y)$.
%
ReI aligns approximations that fit data distributions with disentanglement constraints by causal identification, which reformulates the learning problem as defined by:
\begin{equation} \label{learning_problem}
	 \L( \th; \x^{(i)}, \y^{(i)}) = 
	    \underbrace{\L_{\ell}(\th; \x^{(i)}, \y^{(i)})}_{\text{Data}} ~ + ~ \lambda ~ \underbrace{\L_{\Reg}(\th; \x^{(i)}, \y^{(i)})}_{\text{ReI}} 
\end{equation}
with $\lambda > 0$ being the regularization strength, $\th \in \Th$ a point of the space of DL models and, $\{\x^{(i)} \in \Rspace^M, \y^{(i)} \in \Rspace^n \}_{i=1}^N $ are the observations.
The first term $\L_{\ell}$ is the likelihood function, while the second $\L_{\Reg}$ corresponding to ReI, aligns the latent representation with disentanglement constraints imposed by causal identification (i.e., data + ReI) of $p(\x|do(\y))$.
ReI is different from the weakly-supervised setting of \cite{mitrovic2020representation, peters2017elements, arjovsky2019invariant} which aim at finding the generating mechanisms by imposing some form of invariance to real or augmented data variability. Or to \cite{bouchacourt2018multi, locatello2020weakly, trauble2021disentangled} imposing invariance to shared generative factors between at least pairs of observations while keeping those detected as varying, free. 
%In addition, distinct from \cite{khemakhem2020variational}, our method relies on causal identification valid under a specified DAG $G$ model rather than a statistical notion of model identification.
One additional property of our work is that the encoder is set to produce latent vectors the same size as the inputs $\x \in \Rspace^M$, in similarity to \cite{ ho2020denoising}. 
%In other words, the latent space $\z$ lives in the same space as the observations (this made without restriction of analysis to latents living in lower dimensional spaces). 
This design choice %made in similarity with denoising diffusion models \cite{ ho2020denoising}, 
is made to avoid dependence on the information bottleneck principle while aiming at yielding representations suitable for interpretation. %In this context also, we would like to highlight that our workflow does not rely in any way on the information bottleneck property.

\subsection{Reformulation of the VAE to impose ReI for Disentanglement} \label{Ssec:VAEreformulation}
The variational inference learning problem of the VAE in \cite{kingma2013auto} optimizes the evidence lower bound (ELBO) to approximate the true posterior $p^*(\z|\x)$ given measurements $\{\x^{(i)}\}^N_{i=1}$. The ELBO can be formulated by two terms: a likelihood term $\L_{\ell}(\th; \x^{(i)}) = \expec_{q(\z|\x^{(i)})} \left [  \log p(\x^{(i)}  | \z)   \right ] $ and a regularizer given by the KL divergence as $\L_{\Reg}(\th; \x^{(i)}) = D_{KL}( q(\z|\x^{(i)}) || p(\z'))$. 
In the standard VAE, the prior $p(\z')$ typically a standard Gaussian, is used on the approximate posterior. 
%The $\th,\ph$ are the parameters of the encoder and decoder models, respectively, and optimized over the training dataset. %Often a regularization strength scalar $\lambda$ in front of the regularization term is included to balance tradeoffs between the likelihood and priors, a parameter utilized by the $\beta$-VAE as the information bottleneck.
%
%\item Collider structure control for collider structure
%\item Reformulate the VAE learning formulation to impose ReI as in Sec.\ref{} to control for collider behavior.
%\item (TODO: Implementation details, specially in the context of Eq.\eqref{collider-control}) 
ReI can reformulate the ELBO in the VAE to impose disentanglement constraints through causal identification. 
%, of the posterior $p(\z|\x,do(\y))$ as in Section \ref{Sec:identification} given the collider DAG $G$. 
The reformulated posterior with controlled dependencies between generating variables for disentanglement given the assumed collider DAG $G$ structure in Fig.\ref{fig:sensor_DAG} is equivalent to:
\begin{equation} \label{posterior}
p(\z|\x, do(\y_c)) = p(\x|\z) \expec_{p(\y_{-c})} \left [ p(\z|\y)\right] / p(\x,\y_c)
\end{equation}
involving the observables $\{\x^{(i)},\y^{(i)}\}_{i=1}^N$. The adjustments in Eq.\eqref{posterior} blocks dependencies between variables $\y_c \rightarrow \z \leftarrow \y_{-c}$ when observing $\x$. The full identification derivation of Eq.\eqref{posterior} is included in the Appendix \ref{appendix:identification}.
% ELBO modification by regularizer.
The corresponding ELBO with ReI regularization imposing disentanglement constraints by causal identification results in the reformulated regularizer given as:
%{\color{red} So far analisis of $p(\z|\x, \hy_c)$ but missing is the link to $p(\z|\x, \hy)$}:
%
\begin{equation} \label{ReI}
	\L_{\Reg}(\th, \ph; \x^{(i)}, \y^{(i)}) = 
	D_{KL} \Bigl (q( \z|\x^{(i)}, \y_c^{(i)}) || \expec_{p(\y_{-c})} \left [ p(\z|\y^{(i)})\right] \Bigr )
\end{equation}
Full-derivation is included in Appendix \ref{appendix:elbo}.
Note that imposing the disentanglement constraints through causal identification affects only the ELBO regularizer. The likelihood function in learning problems remains, without modification in general. %This is also consistent with the causal literature \cite{spirtes1996using, pearl2009causality} where free parameters abide to the likelihood function (e.g., least squares) while the adjustments imposed by identification constrain it (e.g., to be zero in the case of linear models \cite{spirtes1995directed}). 
Given these characteristics, we term our method ReI; as the disentanglement constraints required for causal identification can be directly imposed as a regularizer. 
%{\color{red} TODO: Include details about how this is implemented in practice}

\subsection{Experiments on Benchmark Datasets}
We include experiments that show and compare the performance of DL models with ReI against state of the art methods on the task of disentanglement. In this task, the compared methods of \cite{kingma2013auto, higgins2016beta, higgins2018towards, kim2018disentangling, chen2018isolating, locatello2020weakly} are evaluated and in addition we also include experiments in the non-idealized setting of correlated generating factors as in %\cite{li2019learning, geirhos2020shortcut}
\cite{trauble2021disentangled,roth2022disentanglement}. 
We use the VAE+ReI described in Section \ref{Ssec:VAEreformulation} with causal identification derived from the collider DAG in Fig.\ref{fig:sensor_DAG}. %Experiments are conducted that compare performance on this task on both the uncorrelated and the correlated training data settings against the listed state of the art methods.
The datasets used are the standard ML benchmarks used for learning disentangled representations: Shapes3D \cite{kim2018disentangling}, dSprites \cite{higgins2016beta} and MPI3D \cite{gondal2019transfer}. 
%These datasets have been also used in the works of \cite{higgins2016beta, higgins2018towards, kim2018disentangling, locatello2020weakly, trauble2021disentangled, roth2022disentanglement}. %We include performance in both cases when factors are independent and when they are correlated. 

%\subsubsection{Visual Inspection of Disentanglement Supports Collider Behavior}
%\begin{itemize}
%    \item Training with a Supervisory Signal Per Generative Factor
%    \item Training with a Missing Supervisory Signal For One Factor
%    \item Training with a Missing Supervisory Signal For Multiple Factors
%    \item Training in the presence of unidentified noise.
%\end{itemize}

\subsubsection{Explicit Regularizations that Control Collider Behavior Through Causal Identification Do Better at Standard Disentangled Metrics}
The metric of disentanglement performance used here is DCI (Disentanglement, Completeness, Informativeness) scores \cite{eastwood2018framework}. DCI has been established as the most widely accepted metrics of disentanglement performance \cite{locatello2019challenging, locatello2020weakly, trauble2021disentangled, roth2022disentanglement}. DCI evaluations are performed in synthetic datasets generated by either independent or by correlated factors. For the later, we use the extensions in \cite{trauble2021disentangled, roth2022disentanglement} to correlate one, two and three generative factor pairs (where applicable) and one to all factors (1-to-all). We report the average metric and standard deviation (in square brackets) computed over 10 seeds and present the results in Tables \ref{tab:dci_dSprites}, \ref{tab:dci_shapes3d} and \ref{tab:dci_mpi3d} for all three datasets. 
\begin{table}[hbt!]
	\caption{DCI-Disentanglement Performance Comparison on \textbf{dSprites} \cite{higgins2016beta}}
	\label{tab:dci_dSprites}
	\centering
	\begin{tabular}{l rrrr}
		\toprule
	    %& \multicolumn{3}{c}{Uncorr.} & \multicolumn{3}{c}{Pairs: 1} & \multicolumn{3}{c}{Pairs: 2} & \multicolumn{3}{c}{1-to-All} \\
	     %Method & D & C & I & D & C & I & D & C & I & D & C& \\
	     Method & Uncorr. & Pairs: 1 & Pairs: 2 & 1-to-All \\
		\midrule
		$\beta$-VAE \cite{higgins2016beta} & 32.3 {\color{gray} [8.7]} & 9.4 {\color{gray}[2.8]} & 7.8 {\color{gray}[2.5]} & 11.3 {\color{gray}[3.9]} \\
		Factor-VAE \cite{kim2018disentangling} & 25.2 {\color{gray}[7.9]} & 13.1 {\color{gray}[6.7]} & 14.1 {\color{gray}[4.2]} & 14.4 {\color{gray}[3.4]} \\
		$\beta$-TCVAE \cite{chen2018isolating} & 31.3 {\color{gray}[5.8]}& 23.9 {\color{gray}[0.9]} & 11.3 {\color{gray}[5.2]} & 20.3 {\color{gray}[6.1]}\\
		Annealed-VAE \cite{burgess2018understanding} & 39.2 {\color{gray}[3.1]} & 14.8 {\color{gray}[2.2]} & 8.7 {\color{gray}[2.4]} & 14.2 {\color{gray}[0.7]} \\
        $\beta$-VAE + HFS \cite{roth2022disentanglement} & 49.2 {\color{gray}[15.1]} & 19.2 {\color{gray}[2.9]} & 17.5 {\color{gray}[12.3]} & 15.9 {\color{gray}[2.9]}\\
		$\beta$-TCVAE + HFS \cite{roth2022disentanglement} & 53.3 {\color{gray}[9.2]} & 26.2 {\color{gray}[3.0]} & 27.5 {\color{gray}[10.9]} & 24.5 {\color{gray}[4.2]} \\
		VAE + ReI & \textbf{87.5} {\color{gray}[7.1]} & \textbf{88.2} {\color{gray}[7.4]} & \textbf{88.1} {\color{gray}[7.2]} & \textbf{89.4} {\color{gray}[6.1]} \\
		\bottomrule
	\end{tabular}
\end{table}
% 
%[ [32.3, 9.4, 7.8, 11.3],[25.2, 13.1, 14.1, 14.4],[31.3, 23.9,  11.3, 20.3],[39.2, 14.8, 8.7, 14.2],[49.2, 19.2, 17.5, 15.9],[53.3, 26.2,27.5, 24.5],[87.5, 88.2, 88.1, 89.4]]
%[ [38.2,11.3,9.5, 13.9],[30.5,17.6,16.9, 16.7], [35.2, 24.5,14.8,23.8],[41.2,16.3,10.3,14.7],[59.4,21.2,25.8, 17.9],[59.5,28.2,34.9,27.3], [92.3,93.2,93.1,93.5]]
%
\begin{table}[hbt!]
	\caption{DCI-Disentanglement Performance Comparison on \textbf{Shapes3D} \cite{kim2018disentangling}}
	\label{tab:dci_shapes3d}
	\centering
        \resizebox{\columnwidth}{!}{%
	\begin{tabular}{l rrrrr}
		\toprule
		%Method & \multicolumn{3}{c}{Uncorr.} & \multicolumn{3}{c}{Pairs: 1} & \multicolumn{3}{c}{Pairs: 2} & \multicolumn{3}{c}{Pairs: 3} & \multicolumn{3}{c}{1-to-All} \\
		Method & Uncorr. & Pairs: 1 & Pairs: 2 & Pairs: 3 & 1-to-All \\
		%Method & D & C & I & D & C & I & D & C & I & D & C& I & D & C & I \\
		\midrule
		$\beta$-VAE \cite{higgins2016beta} & 70.3 {\color{gray}[9.2]} & 71.2 {\color{gray}[8.9]} & 51.6 {\color{gray}[9.0]} & 36.5 {\color{gray}[4.9]} & 36.3 {\color{gray}[2.7]} \\
		Factor-VAE \cite{kim2018disentangling} & 62.3 {\color{gray}[13.6]}  & 70.8 {\color{gray}[1.6]}  &  58.7 {\color{gray}[5.5]} & 46.1 {\color{gray}[6.1]} & 31.9 {\color{gray}[6.2]} \\
		$\beta$-TCVAE \cite{chen2018isolating} & 77.4 {\color{gray}[3.1]} &  70.2 {\color{gray}[5.6]} & 63.4 {\color{gray}[4.7]} & 38.8 {\color{gray}[11.4]}  & 51.9 {\color{gray}[7.5]} \\
		Annealed-VAE \cite{burgess2018understanding} & 62.1 {\color{gray}[2.6]} & 55.7 {\color{gray}[7.3]} & 30.8 {\color{gray}[6.1]} & 36.2 {\color{gray}[5.2]} & 23.1 {\color{gray}[4.3]} \\
		$\beta$-VAE + HFS \cite{roth2022disentanglement} & 91.8 {\color{gray}[17.9]} & 79.8 {\color{gray}[3.7]} & 67.3 {\color{gray}[5.1]} & 48.7 {\color{gray}[5.0]} & 63.4 {\color{gray}[3.2]} \\
  	$\beta$-TCVAE + HFS \cite{roth2022disentanglement} & 86.3 {\color{gray}[3.6]} & 75.6 {\color{gray}[2.6]} & 66.3 {\color{gray}[7.7]} & 51.7 {\color{gray}[3.8]} & 61.4 {\color{gray}[7.9]} \\
		VAE + ReI & \textbf{95.9} {\color{gray}[5.4]} &  \textbf{96.6} {\color{gray}[3.4]} & \textbf{96.3} {\color{gray}[1.9]} &  \textbf{96.1} {\color{gray}[2.8]} & \textbf{95.8} {\color{gray}[6.3]} \\
		\bottomrule
	\end{tabular}
         }
\end{table}
%
%[ [70.3, 71.2, 51.6, 36.5, 36.3],[ 62.3, 70.8,  58.7, 46.1, 31.9],[ 77.4,  70.2, 63.4, 38.8, 51.9],[ 62.1, 55.7, 30.8, 36.2, 23.1],[ 91.8, 79.8, 67.3, 48.7, 63.4], [86.3, 75.6, 66.3, 51.7, 61.4], [95.9,96.6,96.3,96.1,95.8] ]
%[ [70.3+6.2, 71.2+6, 51.6+6.1, 36.5+3.3, 36.3+1.8],[ 62.3+9.2, 70.8+1.1,  58.7+3.7, 46.1+4.1, 31.9+4.2],[ 77.4+2.1,  70.2+3.8, 63.4+3.2, 38.8+7.68, 51.9+5.05],[ 62.1+1.75, 55.7+4.9, 30.8+4.1, 36.2+3.5, 23.1+2.9],[ 91.8+12.1, 79.8+2.5, 67.3+3.45, 48.7+3.4, 63.4+2.15], [86.3+2.4, 75.6+1.75, 66.3+5.2, 51.7+2.55, 61.4+5.3], [95.9+3.65,96.6+2.31,96.3+1.28,96.1+1.91,95.8+4.21] ]
%
\begin{table}[hbt!]
	\caption{DCI-Disentanglement Performance Comparison on \textbf{MPI3D} \cite{gondal2019transfer}}
	\label{tab:dci_mpi3d}
	\centering
        \resizebox{\columnwidth}{!}{%
	\begin{tabular}{lrrrrr}
		\toprule
		%& \multicolumn{3}{c}{Uncorr.} & \multicolumn{3}{c}{Pairs: 1} & \multicolumn{3}{c}{Pairs: 2} & \multicolumn{3}{c}{Pairs: 3} & \multicolumn{3}{c}{1-to-All} \\
		%Method & D & C & I & D & C & I & D & C & I & D & C& I & D & C & I \\
		Method & Uncorr. & Pairs: 1 & Pairs: 2 & Pairs: 3 & 1-to-All \\
		\midrule
		$\beta$-VAE \cite{higgins2016beta} & 25.9 {\color{gray}[7.9]} & 18.3 {\color{gray}[2.4]} & 23.7 {\color{gray}[1.3]}& 11.3 {\color{gray}[0.5]} & 11.2 {\color{gray}[2.0]}\\
		Factor-VAE \cite{kim2018disentangling} & 26.6 {\color{gray}[2.0]} & 22.8 {\color{gray}[2.8]}& 28.2 {\color{gray}[1.5]} & 11.0 {\color{gray}[0.9]} & 13.8 {\color{gray}[0.8]} \\
		$\beta$-TCVAE \cite{chen2018isolating} &27.3 {\color{gray}[1.0]}& 20.9 {\color{gray}[0.7]} & 22.8 {\color{gray}[1.4]} & 11.1 {\color{gray}[1.7]} & 14.5 {\color{gray}[1.5]} \\
		Annealed-VAE \cite{burgess2018understanding} & 11.4 {\color{gray}[1.3]} & 12.3 {\color{gray}[1.9]} & 11.9 {\color{gray}[0.4]} & 10.7 {\color{gray}[1.2]} & 13.1 {\color{gray}[0.8]} \\
		$\beta$-VAE + HFS \cite{roth2022disentanglement} & 32.9 {\color{gray}[3.2]} & 29.2 {\color{gray}[2.2]} & 27.3 {\color{gray}[0.6]} & 13.8 {\color{gray}[1.3]} & 15.7 {\color{gray}[1.2]} \\
  	$\beta$-TCVAE + HFS \cite{roth2022disentanglement} & 32.6 {\color{gray}[3.4]} & 28.6 {\color{gray}[4.1]} & 29.1 {\color{gray}[0.7]} & 11.4 {\color{gray}[ 3.9]} & 15.2 {\color{gray}[1.3]} \\
		VAE + ReI & \textbf{73.5} {\color{gray}[5.5]} & \textbf{72.6} {\color{gray}[7.2]} & \textbf{74.3} {\color{gray}[6.3]} & \textbf{71.9} {\color{gray}[3.2]} & \textbf{73.5} {\color{gray}[4.5]}\\
		\bottomrule
	\end{tabular}
         }
\end{table}
%
%[ [25.9, 18.3, 23.7, 11.3, 11.2],[ 26.6, 22.8, 28.2, 11.0, 13.8], [27.3, 20.9, 22.8, 11.1, 14.5],[ 11.4, 12.3, 11.9, 10.7, 13.1],[32.9, 29.2, 27.3, 13.8, 15.7],[ 32.6, 28.6, 29.1, 11.4, 15.2],[73.5,72.6,74.3,71.9,73.5] ]
%[ [25.9+5.35, 18.3+1.6, 23.7+0.85, 11.3+0.3, 11.2+1.35],[ 26.6+1.35, 22.8+1.9, 28.2+1.0, 11.0+0.6, 13.8+.55], [27.3+0.7, 20.9+0.45, 22.8+0.95, 11.1+1.15, 14.5+1.0],[ 11.4+0.9, 12.3+1.25, 11.9+0.25, 10.7+0.8, 13.1+0.55],[32.9+2.15, 29.2+1.5, 27.3+0.4, 13.8+0.85, 15.7+0.8],[ 32.6+2.15, 28.6+1.5, 29.1+0.45, 11.4+0.85, 15.2+0.89],[73.5+3.7,72.6+6.1,74.3+4.3,71.9+4.6,73.5+5.1] ]

%In tables \ref{tab:dci_dSprites}, \ref{tab:dci_shapes3d} and \ref{tab:dci_mpi3d} 
Here, we see that the DCI performance degrades throughout all datasets as the number of correlated pairs increases for most of the methods compared. These do not seem to be well equipped to handle an increasing number of correlated generating factors. %This increase in number and severity of correlated factors has a severe negative effect on their disentanglement performance. 
The observed degradation can be explained by the behavior of a collider where factors without control introduce dependencies between them producing entanglements. The severity of such dependencies depends on the number of factors that remain unadjusted for. The fact that the compared methods do not address this collider behavior explicitly, explains the lower performance as the number of correlated pairs increases. By explicitly addressing this type of bias by leveraging the power of causal models, the performance of ReI remains more or less invariant to the number of correlation pairs and their strength, as long as causal identification is possible. This is one of the main benefits of ReI, which of course comes at the cost of requiring a supervisory signal (labels in these cases) to identify the effects of the generating factors.

In contrast, the MPI3D dataset comes from real images captured from a moving robotic arm. The relatively lower performance of all methods on this dataset can be attributed to the fact that it comes from a real-world scenario with several unmeasured factors of variation. The images in MPI3D are obtained from three different cameras each affected by sensor noise, blur, illumination changes from view. The unconstrained setting, results in representations entangled from conditional dependencies with such variables $\u_{\x}$ in Fig.\ref{fig:sensor_DAG}. In this sense, the collider DAG structure offers an explanation and the conditions for control. Control can be performed through additional experiments, gathering additional observations about $\u_{\x}$,or assuming a parametric form to provide supervision. This, is an advantage over the methods compared, where explanations and ways to remediate, remain a black-box.

\section{Experiments on Real-world Dataset}
\label{Sec:experimentation } 

Experiments were %conducted that test the ability of the proposed causal-guided regularization approach on learning representations that explain the data while simultaneously being capable of producing representations in disentangled form. 
%As described in Section \ref{Sec:supervised}, we considered a special case of learning representations under soft-supervision where for each measurement $\x \in \Rspace^{N_{\x}}$ there is a set of class membership scores $\y \in \Rspace^{N_{\y}}$. 
%Validation was 
conducted in spectroscopic applications, specifically using data from a laser induced breakdown spectroscopy (LIBS) instrument. 
%We opted for this alternative given the rigorous investigations that have been made and that have produced at least at a coarse level well established physics principles explaining the data formation. Known qualititative characterizations of the spectral response effects of certain physical mechanisms in the corresponding measurements will be used for comparison against those learned. 
LIBS is a remote sensing technology used to predict the chemical composition of geomaterials (e.g., rocks, soil) based on its signatures. On Mars, the ChemCam LIBS based instrument
%s are capable of investigating $<1$mm size samples by laser excited samples from up to 7m distances. It 
is equipped with a 1064nm laser and ultraviolet, visible and near infrared band spectrometers; which altogether is capable of collecting the sample's spectral signatures between 240-905nm. %It is through the analysis of such signatures, that the chemical composition of a sample can be extracted.
Focus here, is applying the ReI framework directed towards: (1) representation disentanglement, (2) prediction and (3) transfer. (1) learns representations characteristic of specific chemical elements. (2) uses the learned representations to predict chemical content, while (3), tests for robustness to dataset shifts by training data from Earth in a controlled setting while deployment is in the wild on Mars. Additional details are included in Appendix \ref{Ssec:data_details}

%\subsection{Dataset}
%The ChemCam LIBS instrument \cite{wiens2012chemcam} datasets contain raw and denoised spectra obtained from a variety of targets (e.g., rocks, soil) and from reference calibration standards of known and certified chemical composition. The specific datasets we employ consists of spectrally resolved LIBS signal measurements collected on Earth in a laboratory setting from a set of $\sim$ 585 reference calibration standards \cite{Clegg:2017} and on Mars from a set of 10 reference standards of known true composition. Each target is repeatedly shot (e.g.,  50 times) following each time measurements of the full 240-905 nm LIBS signal. %This process is repeated at multiple locations for each target which amounts to $\sim$41,000 uncleaned LIBS signals. 
%After collection, wavelengths within the bands [240.811,246.635], [338.457,340.797], [382.13,387.859], [473.184,492.427], [849,905.574] were ignored out consistent with practices of the ChemCam team \cite{Clegg:2017}. 

\begin{figure*}[t]
	\centering
	\begin{subfigure}[b]{0.86\textwidth}
		\centering
		\includegraphics[width=\textwidth]{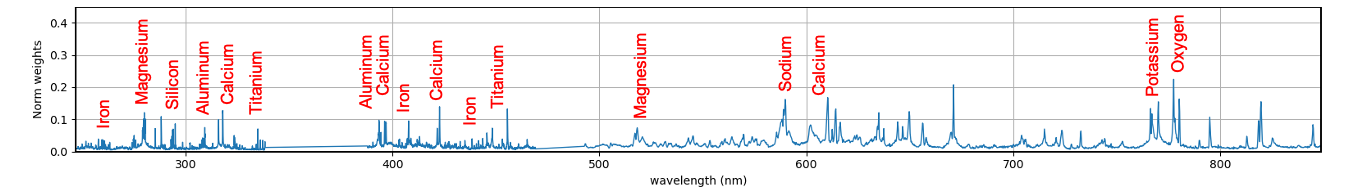}
		\caption{\scriptsize{Standard VAE learns representations with entangled effects from other chemical elements.}}
		\label{fig:rep_unblocked}% label for this sub-figure
	\end{subfigure} % space out the images a bit
	\begin{subfigure}[b]{0.12\textwidth}
		\centering
		\includegraphics[width=\textwidth]{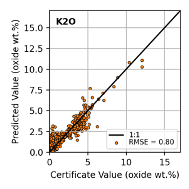}
		\caption{\scriptsize{Predictions}}
		\label{fig:reg_unblocked}% label for this sub-figure
	\end{subfigure}
	
	\begin{subfigure}[b]{0.86\textwidth}
		\centering
		\includegraphics[width=\textwidth]{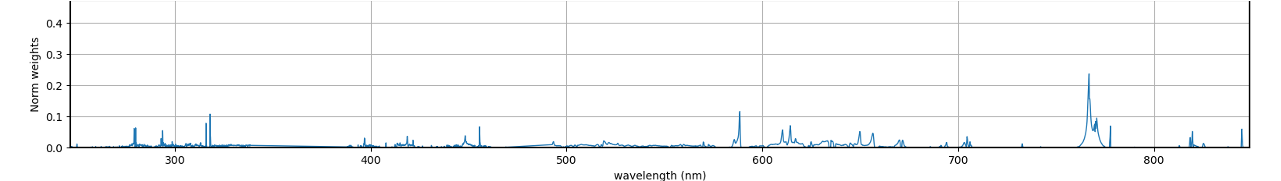}
		\caption{\scriptsize{VAE+ReI blocking generating factors from all chemical elements except from sensor noise.}}
		\label{fig:rep_blocked_noisy}% label for this sub-figure
	\end{subfigure} 
	\begin{subfigure}[b]{0.12\textwidth}
		\centering
		\includegraphics[width=\textwidth]{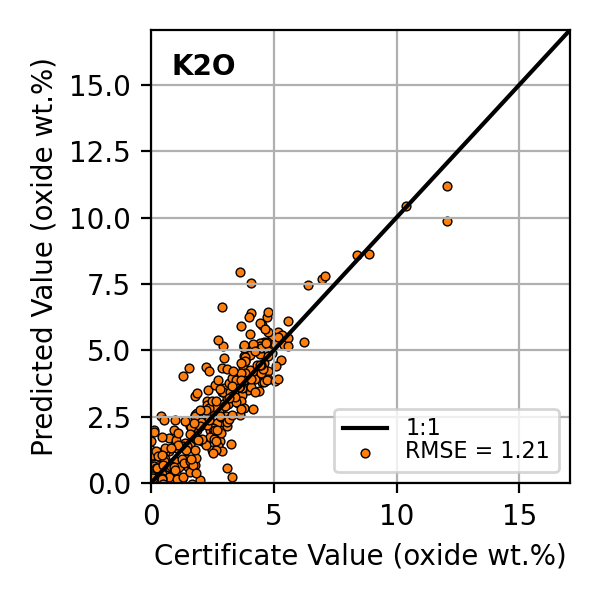}
		\caption{\scriptsize{Predictions}}
		\label{fig:reg_blocked_noisy}% label for this sub-figure
	\end{subfigure}

	\begin{subfigure}[b]{0.86\textwidth}
	   \centering
	   \includegraphics[width=\textwidth]{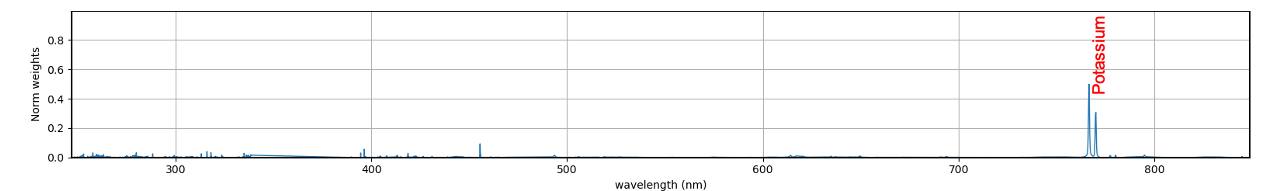}
	   \caption{\scriptsize{VAE+ReI blocking all generating variables.}}
	   \label{fig:rep_blocked}% label for this sub-figure
    \end{subfigure} 
    \begin{subfigure}[b]{0.12\textwidth}
	   \centering
	   \includegraphics[width=\textwidth]{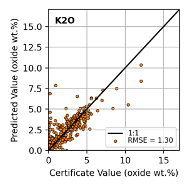}
	   \caption{\scriptsize{Predictions}}
	   \label{fig:reg_blocked}% label for this sub-figure
    \end{subfigure}
	\caption{Comparison of the learned representations for chemical oxide K$_2$O.}
	\label{fig:chemcam_rep}
\end{figure*}

\subsection{Representation disentanglement} \label{Ssec:exp_representations}
%La calibracion que como ya antes fue mencionado consiste en el aprendizaje de representaciones que caracterizen el contenido quimico dada las mediciones de un material sampleado es probada en esta sub-seccion. Para esto, Example pairs of the LIBS signal and ground truth chemical composition are utilized from the reference standard targets to learn the dictionary elements. El modelo lineal descrito en Eq. \eqref{backward} es utilizado para esta tarea. Para evaluar el desempenio, comparamos qualitativamente las representaciones obtenidas con nuestro metodo versus el metodo convencional utilizado en standard deep learning based approaches de utilizar unicamente el input y output y dejar que la red neuronal aprenda sus parametros todos de manera libre.

The abilities of ReI for learning disentangled representations were evaluated here. Training utilizes example pairs $\{\x^{(i)}, \y^{(i)}\}_{i=i}^N$ of LIBS signal $\x \in \Rspace^{5485}$ measurements and corresponding true chemical composition $\y$. 
% DAG
%
%\begin{figure} [h]
%	\centering 
%	\includegraphics[width=0.75\linewidth]{figures/DAG_chemcam2}%multiview_side_2}}
%	\caption{DAG for ChemCam.}
%	\label{fig:DAG_chemcam} 
%\end{figure}
%
%The assumed data generation process of the instrument for this task is similar to the DAG of Fig.\ref{fig:sensor_DAG}. Main difference in this case however, is that there is an active illumination source (i.e., a laser) $U_0$ as a parent of all the $\y_i$'s in Fig.\ref{fig:sensor_DAG} that interacts with a target material $Y \in \Rspace^{n}$ composed of a 
Percentages $\y_c$ represent $\%$ oxide composition for $c  \in \{1,...,11\}$ indexing $ \{ \text{SiO}_2, \text{TiO}, \text{Al}_2\text{O}_3,$ $\text{FeO}_\text{T}, \text{MgO}, \text{MnO}, \text{CaO}, \text{Na}_2\text{O}, \text{K}_2\text{O}, \text{CO}_2, \text{H}_2\text{O}\}$ and sensor noise is $\u_{\x} \in \Rspace^{5485}$. %Unmeasured pulse-target oxide interactions are denoted by $U_1,...,U_n$ with $U_i \in \Rspace^N  \quad \forall i$ while $U_x \in \Rspace^N$ represents spectrally resolved noise. %The problem is to learn a representation that predicts the causal effect of $Y_i$ given $X$ from training data. We make use of instrumental variables to learn representations of the unmeasured pulse-target interaction components $U_1,...,U_n$ and denote these by $\th_1,...,\th_n$ with $\th_i \in \Rspace^N \quad \forall i$. The variables $S_{Y_i} \text{ for } i \in \{1,..,N\}$, $S_{U_E}$ and $S_{U_X}$ are unmeasured noise sources with infuence on the variables pointed by their corresponding directed edge connections.
%
%Prediction performance of the chemical composition from the learned representations was evaluated in this sub-Section. 
%For this, 
%{\color{red} primero evaluamos de manera cualitativa y cuantitativa las representaciones producidas por el metodo de regularizacion causal propuesto aqui a la luz de la respuesta espectral que es conocida por caracterizar estos oxidos quimicos. Para esto, usamos el metodo mas simple en donde entrenamos una red neuronal que consiste unicamente de una capa lineal y comparamos las representaciones producidas en los dos casos: cuando estas son restrinjidas bajo la regularizacion causal y cuando no son restrinjidas.  }
Qualitative evaluations of the representations from ReI derived from the collider DAG in Fig.\ref{fig:sensor_DAG} were performed in light of the known characteristic spectral response of each chemical oxide. We used an MLP architecture and compared the representations in three cases: (1) the standard VAE,  (2) VAE+ReI with all factors identified except for sensor noise and (3) ReI with all generating factors identified. Training used $585$ reference targets under leave one out while testing was done on the target left out until all are covered. Additional implementation details are included in Appendix \ref{Ssec:implementation_details}.
A representative example on the learned representations corresponding to chemical oxide K$_2$O is shown in Fig.\ref{fig:chemcam_rep}. These were generated by sampling from $q(\z|\x^{(i)}, \y^{(i)}_c)$ with $\y_c$ as composition of K$_2$O and averaging over $L=100$ samples. %Here, we illustrate some of the qualitative and quantitative capabilities of the causally constrained method proposed here compared to the unconstrained one. 
%In the former, we show the learned representation corresponding to specific chemical elements of interest. In the later, we show the results of prediction versus ground truth plots and their corresponding overall root mean squared error (RMSE). For this purpose, the $\sim400$ reference calibration standard targets, are used for training under leave one out while testing on data from the standard left out until all reference standards are covered. 
Figs.\ref{fig:rep_unblocked}, \ref{fig:rep_blocked_noisy} and \ref{fig:rep_blocked} shows the learned representations for K$_2$O in: (1) VAE, (2) VAE+ReI with sensor noise $\u_{\x}$ uncontrolled and (3) VAE+ReI with control for all generating factors. The vertical axis of each plot shows the normalized magnitude and the horizontal axis represents spectral wavelength importance. Figs.\ref{fig:reg_unblocked}, \ref{fig:reg_blocked_noisy} and \ref{fig:reg_blocked} illustrate the corresponding prediction performance $\ty_c$ of the three cases using the representations $\z$ along with a trained linear prediction head. Prediction performance by looking into point distribution along the 1:1 line and as measured by the root mean squared error (RMSE) shows similar performances in all three cases; with a marginal advantage of the standard VAE (i.e., (1) 0.8, (2) 1.21, (3) 1.30). However, these all come from distinct learned representations with key observations supporting evidence of collider behavior. First, note that K$_2$O (Potassium oxide) is known and expected to respond to wavelengths around $\sim770$nm as labeled in Fig.\ref{fig:rep_unblocked}, illustrating the ground truth expected spectral responses of a variety of chemical elements. 
The standard VAE in Fig.\ref{fig:rep_unblocked} resulted in a representation with spectral peaks deemed important spread throughout the entire spectrum. This is indicative of conditional dependencies between the generative factors $\y_{-c} = \{\y_j: j \neq c\}$ and K$_2$O through the path $\y_{-c} \rightarrow \x \leftarrow \y_c$.
Fig.\ref{fig:rep_blocked_noisy} in contrast shows the resulting representation obtained by VAE+ReI with control for dependencies between the generative factors except for those from sensor noise $\u_{\x}$. Although most of the wavelengths previously deemed important were flattened, some small spectral peak patterns from Fig.\ref{fig:rep_unblocked} persisted. This, due to conditional associations between the paths $\y_{-c} \rightarrow \x \leftarrow \u_{\x}$ and $\u_{\x} \rightarrow \x \leftarrow \y_c$. 
Finally, Fig.\ref{fig:rep_blocked} illustrates the representation by VAE+ReI with control for all generative factors. Most wavelengths were brought down to zero except for the two strong peaks at $\sim770$nm in alignment with the expected spectral response for K$_2$O. Identification thus produced representations well aligned with the expected effects of the generating factors. 

The empirical evidence provided supports our claim that standard generative representation learning models ill-suffer from collider bias. Note that downstream tasks, such as the prediction of in-distribution examples and without visualizations of the learned representations as exemplified by Figs.\ref{fig:reg_unblocked}, \ref{fig:reg_blocked_noisy} and \ref{fig:reg_blocked} can obscure the aforementioned illness. However, illustrations of the representations of the effects of generating factors clearly shows evidence of this problem, with effects supporting collider behavior. These findings, thus provide a plausible alternative explanation to the spurious association problems between factors found in \cite{geirhos2020shortcut, pezeshki2021gradient}, 
%\cite{glocker2019machine, geirhos2020shortcut, pezeshki2021gradient, banerjee2021reading}, 
to fairness \cite{zhao2017men}, and provide a venue for analysis and remediation through causality as viewed by \cite{Pearl:2010} and tackled here by ReI. We would like to note also that the learned representations from the VAE+ReI are amenable for interpretation while also explain the effects of generating factors as they relate to the effects of the measuring apparatus, supporting understanding.

\subsection{Prediction and Transfer}

\begin{figure*}[t]
	\centering
	\begin{subfigure}[b]{0.20\textwidth}
		\centering
		\includegraphics[width=\textwidth]{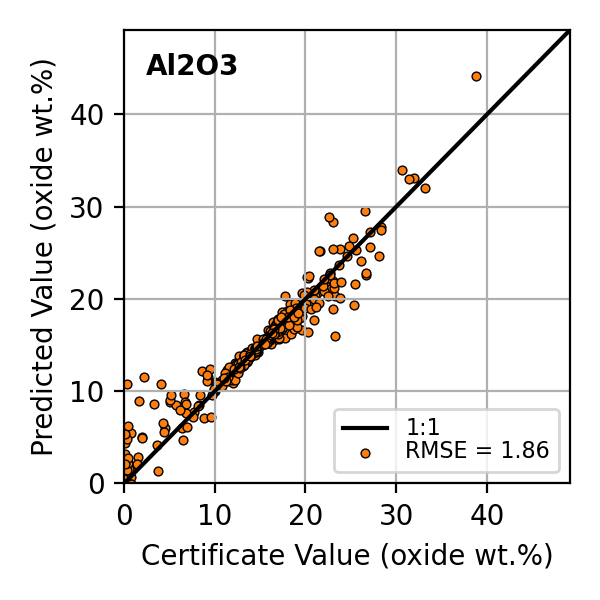}
		\caption{Standard: ID}
		\label{fig:al2o3_in_conv}% label for this sub-figure
	\end{subfigure} 
	\begin{subfigure}[b]{0.20\textwidth}
		\centering
		\includegraphics[width=\textwidth]{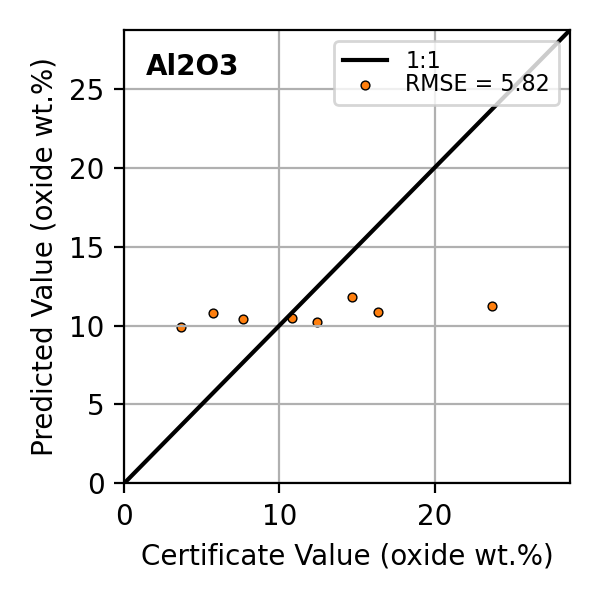}
		\caption{Standard: OOD}
		\label{fig:al2o3_i}% label for this sub-figure
	\end{subfigure}
	\begin{subfigure}[b]{0.20\textwidth}
		\centering
		\includegraphics[width=\textwidth]{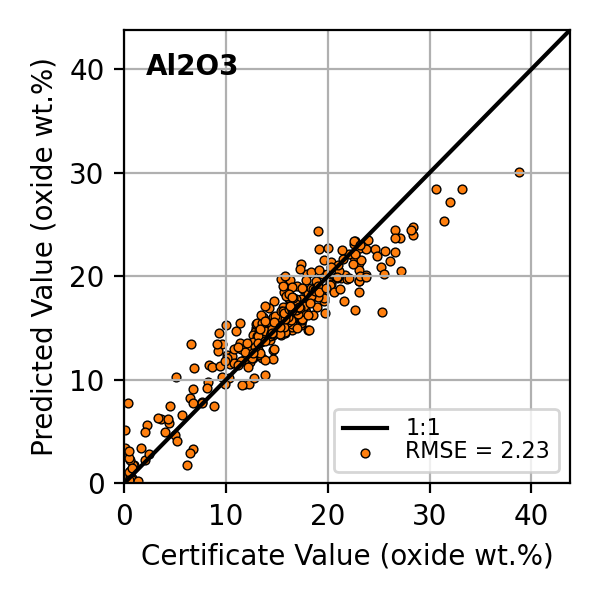}
		\caption{ReI:ID}
		\label{fig:al2o3_in_causal}
	\end{subfigure} 
	\begin{subfigure}[b]{0.20\textwidth}
		\centering
		\includegraphics[width=\textwidth]{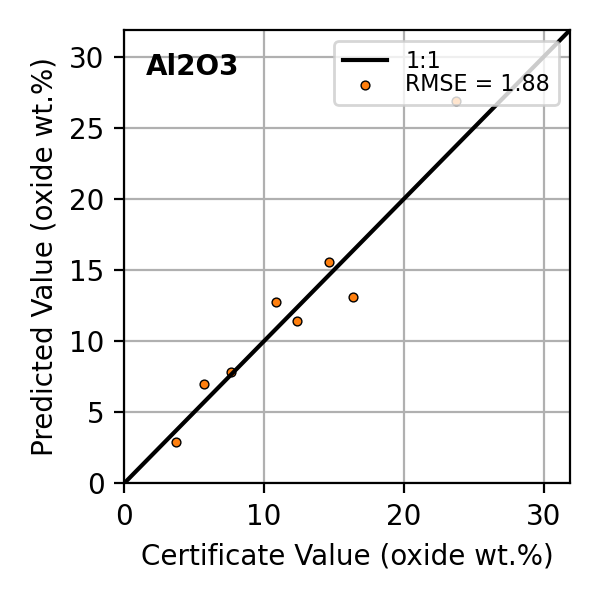}
		\caption{ReI: OOD}
		\label{fig:al2o3_o}% label for this sub-figure	
	\end{subfigure}
	
	\begin{subfigure}[b]{0.20\textwidth}
		\centering
		\includegraphics[width=\textwidth]{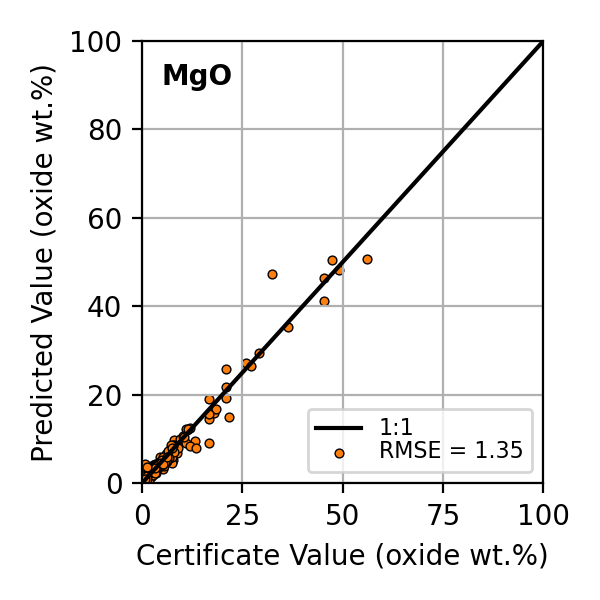}
		\caption{Standard: ID}
		\label{fig:mgo_in_conv}% label for this sub-figure
	\end{subfigure} 
	\begin{subfigure}[b]{0.20\textwidth}
		\centering
		\includegraphics[width=\textwidth]{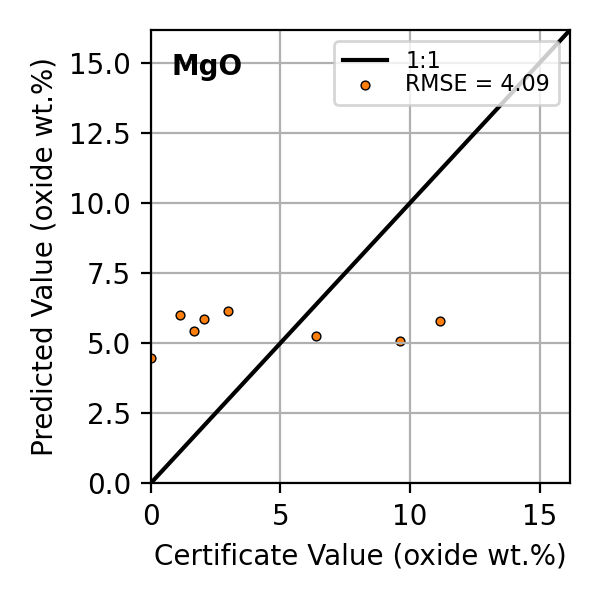}
		\caption{Standard: OOD}
		\label{fig:mgo_i}% label for this sub-figure
	\end{subfigure}
	\begin{subfigure}[b]{0.20\textwidth}
		\centering
		\includegraphics[width=\textwidth]{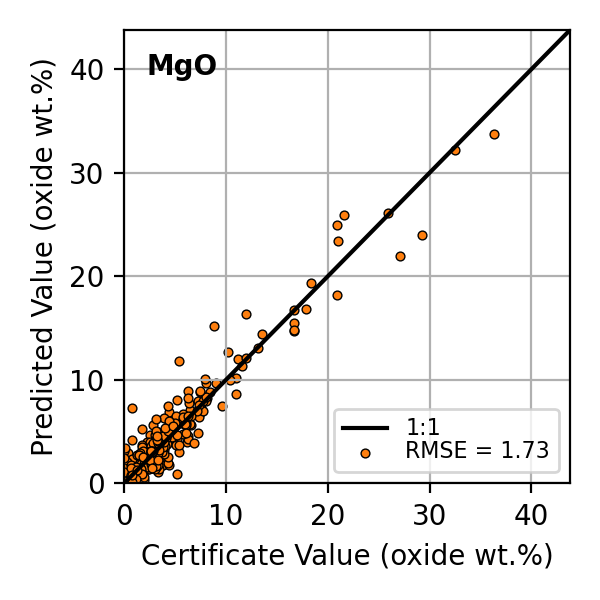}
		\caption{ReI: ID}
		\label{fig:mgo_in_causal}
	\end{subfigure} 
	\begin{subfigure}[b]{0.20\textwidth}
		\centering
		\includegraphics[width=\textwidth]{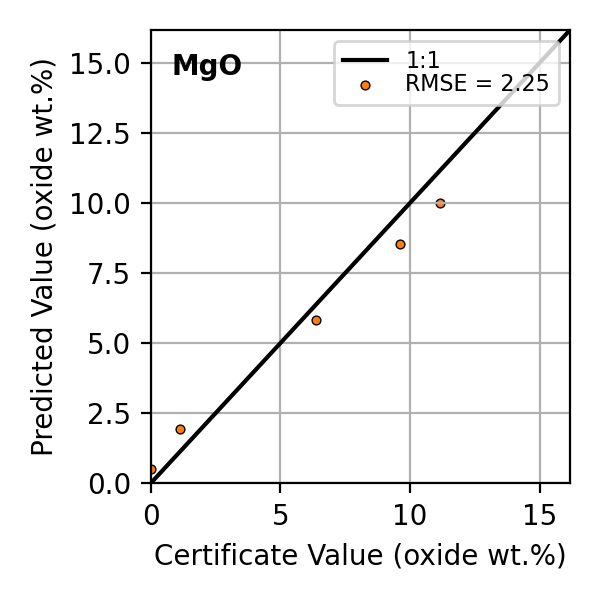}
		\caption{ReI: OOD}
		\label{fig:mgo_o}
	\end{subfigure}
	\caption{Performance comparison of the learned representations in out-of-distribution transfer.}
	\label{fig:transfer}
\end{figure*}
%Calibration training uses a dataset collected under carefully controlled settings (e.g., sensor-to-target distance, temperature, laser angle of incidence) in a lab on Earth while deployment occurs in the wild on Mars. The work here follows that of the calibration task however the difference here is that we jointly leverage three data generation process models from two sources and the target domain: (1) calibration data $C$ collected on Earth under settings $A$ with sensor $K_1$, (2) calibration data $C$ collected on Mars under settings $B$ with sensor $K_2$ and (3) deploying on Mars under sensor settings $C$ and sensor $K_2$. We use transportability \cite{pearl2011transportability} (i.e., transfer learning) of invariant factor effects between domains while adjusting for those that differ through analysis of the DAGs. Similar to \ref{fig:chemcam_rep} evaluations will be conducted by looking at spectra versus E3M response and its predictive accuracies under out-of-distribution inputs from Earth and Mars targets of known composition.

%In the previous experimentation, we investigated the performance of the learned representations both qualitative and quantitative. Qualitatively, it was observed that the learned dictionary elements align well with the expectated spectral response of chemical composition according to physics principles. Quantitatively, a small difference between the causal and standard methods was observed, where the causal achieved a marginally lower performance. 
Quantitative comparisons on the robustness to dataset shifts is performed here. Dataset shifts originate here by training from data collections of LIBS from targets on Earth in a laboratory setting while deployment occurs in the wild on Mars. \cite{Clegg:2017} found that the Martian environment has effects that shift the distribution of measurements relative to Earth. 
%and provided a manual engineered approach for their correction. 
This task, then seeks to investigate the transferability of the learned representations in the presence of OOD shifts.

%For this, data from reference calibration standards acquired with a LIBS sensor but this one deployed in Mars. In like manner of the sensor, also the reference calibration standards are identical copies (with minor differences potentially caused by fabrication nonuniformities). 
%The learned representations from Earth which have not ever seen data coming from Mars are evaluated under this context. 
Example representative results are included in Fig.\ref{fig:transfer} which shows true versus prediction plots for two element oxides Al$_2$O$_3$ and MgO. Four leftmost Figs.\ref{fig:al2o3_in_conv},\ref{fig:mgo_in_conv},\ref{fig:al2o3_i},\ref{fig:mgo_i} corresponds to performance results from the standard VAE+FC linear head whereas the rightmost four Fig.\ref{fig:al2o3_in_causal},\ref{fig:mgo_in_causal},\ref{fig:al2o3_o},\ref{fig:mgo_o} shows those from VAE+ReI+FC. Figs. \ref{fig:al2o3_in_conv},\ref{fig:mgo_in_conv},\ref{fig:al2o3_in_causal},\ref{fig:mgo_in_causal} show that both VAE and VAE+ReI present similar performance for in-distribution example testing (under leave one out), with the VAE being marginally better in terms of RMSE. In contrast, Figs.\ref{fig:al2o3_i},\ref{fig:mgo_i},\ref{fig:al2o3_in_causal},\ref{fig:mgo_in_causal} shows significant differences in performance in the OOD cases. VAE+ReI presents better behaved performance and outperforms by larger margins compared to the VAE.
%For reference, we've also included Figs.\ref{fig:al2o3_in_conv},\ref{fig:mgo_in_conv} and \ref{fig:al2o3_in_causal},\ref{fig:mgo_in_causal} which show accuracy in examples without distribution shifts under leave one out. 
%
Note that even though the VAE presents an advantage over VAE+ReI for in-distribution performance, that this is not the case for OOD examples. The disentanglement provided by VAE+ReI shows better robustness against OOD examples. This behavior, is consistent with findings by \cite{tsipras2018robustness}, where highly predictive non-robust features in the data tend to reduce learner performance when presented with OOD examples. 
							% Experimentation
\section{Conclusions} 
\label{Sec:conclusions}

%En este estudio, tratamos el problema de representacion de datos por medio de el aprendizaje de elementos de diccionario desde la lente de causalidad. Analizamos una grafica que representa un modelo de generacion de deatos asumido y utilizando las reglas del do-calculus identificamos la causa de efectos condicional implicada por dicho modelo de generacion de datos y aprendizaje de elementos de diccionario para realizar predicciones. Hicimos la coneccion de que en el caso de funciones lineales la identification de causa de efectos corresponde a imponer restricciones de ortogonalidad. Las cuales encontramos resultan en una descomposicion que corresponde a disjoint union of subspaces. Validamos nuestros hallazgos en la aplicacion de prediccion de composicion quimica de materiales utilizando mediciones de un laser induced breakdown spectroscopy sensor. En este encontramos que los elementos de diccionario aprendidos corresponden a las spectral response signatures que caracterizan cada elemento quimico. Estas encontramos son robustas en la presencia de out-of-distribution examples en el caso cuando son aprendidas en laboratorio en la tierra y deployed in Mars.

%En esta investigacion proponemos el uso de regularizaciones que imponen restricciones causales para el problema de representation learning. Estas regularizaciones son derivadas por medio del control del libre flujo de informacion entre variables y estan al servicio de la identificacion de causa y efecto. 
In this work, we proposed ReI: a regularization method that aligns DL models to domain knowledge by leveraging the DAG. We argued that standard disentangled learning models are ill-biased by collider behaviour and showed supporting empirical evidence of this. In a variational framework, we showed how analysis of the DAG under the lens of causality can be used to control for collider bias via ReI in representation learning problems. Empirical evidence shows ReI is capable of learning the effects between the generating factors and the sensor, removing collider bias, producing representations in disentangled form, generalizable to OOD example cases and supporting interpretation of both factor effects and manipulations of these for sampling posterior generation.

							% Conclusions

%{\small
%	\bibliographystyle{files_iclr/iclr2024_conference}
%	\bibliography{files_iclr/refs}
%}

\appendix
\section{Background}
\label{appendix:background} 

% Idea: Causal Dictionary Learning operating under a supervised learning strategy. We seek to tie in or link the causal effect of physical variable with a dictionary element or  atom.

\subsection{The rules of $do$-calculus} \label{Ssec:axioms}
The axioms branded under the $do$-calculus are presented here.

In terms of notation, in a DAG $G$, $G_{\overline{X}}$ and $G_{\underline{X}}$  denote, respectively, the graphs obtained by deleting the incoming and outgoing arrows at node $X$. 

The rules of interventional $do$-calculus according to Pearl are given as \cite{pearl1995causal, Pearl:2010}:

\noindent "\textbf{Rule 1} (Insertion/deletion of observations):

$p(y | \hat{x}, z,w) = p(y | \hat{x},w)$      \hspace{1em} if \hspace{0.5em} $(Y \indep Z) | X,W)_{G_{\overline{X}}}$

\noindent \textbf{Rule 2} (Action/observation exchange):

$p(y | \hat{x}, \hat{z},w) = p(y | \hat{x},z, w)$      \hspace{0.5em} if \hspace{0.5em} $(Y \indep Z) | X,W)_{G_{\overline{X} \underline{Z}}}$

\noindent \textbf{Rule 3} (Insertion/deletion of actions):

$p(y | \hat{x}, \hat{z},w) = p(y | \hat{x},w)$      \hspace{1em} if \hspace{0.5em} $(Y \indep Z) | X,W)_{G_{\overline{X Z}} } $"

These graphical rules encompass the foundational principles of the $do$-calculus \cite{pearl1995causal}. By analyzing the DAG and employing these rules, it becomes possible to characterize the effects of interventions $do(x)$ in terms of ordinary probability distributions of observations. This process, known as identification in the context of causal inference, serves as the primary analytical tool for elevating relationships between variables from mere correlation to causation.

\section{Causal Effect Identification in Collider Structure}
\label{appendix:identification} 

\begin{figure}[h]
	\centering
	\begin{subfigure}[b]{0.30\textwidth}
		\centering
		\includegraphics[width=0.5\textwidth]{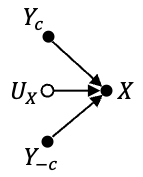}
		\caption{DAG $G$, $G_{\overline{\y_c}}$ and $G_{\overline{\y_{-c}}}$}
		\label{fig:g}% label for this sub-figure
	\end{subfigure} % space out the images a bit
	\begin{subfigure}[b]{0.30\textwidth}
		\centering
		\includegraphics[width=0.5\textwidth]{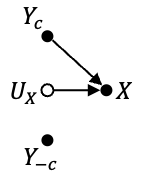}
		\caption{DAG $G_{\underline{\y_{-c}}}$}
		\label{fig:g_yi_}% label for this sub-figure
	\end{subfigure}
	\begin{subfigure}[b]{0.30\textwidth}
		\centering
		\includegraphics[width=0.5\textwidth]{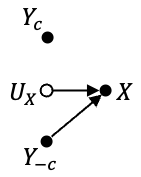}
		\caption{DAG $G_{\underline{\y_c}}$.}
		\label{fig:g__yi}% label for this sub-figure
	\end{subfigure} 
	\caption{Collider DAG}
	\label{fig:dag_g}
\end{figure}
Identification of the causal effects $p(\x|\hy_c)$ in Eq.(1) involves application of the rules of the $do$-calculus by leveraging the causal assumptions encoded in the DAG. This is used to convert probabilities of interventions to expressions involving only ordinary probabilities of observations. %The DAG, simplified to only three generative factors is shown in \ref{fig:dag_g} with $\y_c, \w_c$ measured and unobserved $\u_x$ (i.e., the sensor noise). 
The DAG in Fig.\ref{fig:g} is a representation of the data generative model of independent factors $p(\y) = \prod_{c =1}^n p(\y_c)$ assumed in \cite{kingma2013auto, higgins2016beta, higgins2018towards, kim2018disentangling} and a simplification of Fig.1. In this case however, with only three factor variables $\y_c, \y_{-c},\u_x$. The DAG structure in Fig.\ref{fig:g} contains a collider at $\x$. A collider is represented in a DAG by a node where two or more arrows or paths converge. A collider, produces conditional associations between the generating factors $\y_c, \y_{-c}, \u_x$, even if they are not causally related. This phenomenon is known as collider bias or selection bias \cite{berkson1946limitations, kim1983computational, pearl2009causality}. This needs to be accounted for, when training a DL model to learn from the joint distribution $p(\x, \y, \u_x)$ over the observations, otherwise it is prone to produce biased models. Again, we do this through derivations involving the effects of interventions $p(\x|\hy_c)$ in the system encoded by Fig.\ref{fig:g}.
% Fig.\ref{fig:g_yi_} removes edges of arrows that point outward and Fig.\ref{fig:g__yi} those that point inward of $\y_c$, respectively. Given that under the DAG $G_{\underline{\y_c}}$, $\z$ and $\y_c$ are not $d$-Separated requires introducing additional conditioning on $\w_c$ to block the open path $\y_c \leftrightarrow \w_c \rightarrow \z$. 
Application of the law of total probability in Eq.\eqref{ltp} and the chain rule in Eq.\eqref{chain} both valid under probabilities of interventions \cite{pearl2009causality} yields:
\begin{eqnarray} 
p(\x | \hy_c)
= \sum_{\y_2,\u_x} p(\x,  \hy_c, \hy_{-c}, \u_x) / p(\hy_c) \label{ltp} \\
 = \sum_{\y_{-c},\u_x} p(\x | \hy_c, \y_{-c}, \u_x) p(\hy_c |\y_{-c}, \u_x) p(\y_{-c},\u_x) / p(\hy_c)  \label{chain} \\
  = \sum_{\y_{-c},\u_x} p(\x | \hy_c, \y_{-c}, \u_x) p(\y_{-c},\u_x)  \label{rm} \\
 =  \sum_{\y_{-c},\u_x} p(\x | \y, \u_x) p(\y_{-c},\u_x) = \expec_{\y_{-c},\u_x} \left [ p(\x|\y, \u_x)\right] \label{ic}
 \end{eqnarray}
Note that we have used for easy of notation $\hy_c = do(\y_c)$. Eq.\eqref{rm} follows by definition of the $do$ operator implying no effect on an intervention conditioned on any variables (i.e., $p(\hy_c|\y_{-c}) = p(\hy_c) = 1$). The later, follows from the definition of an intervention, where there is no uncertainty on the variable being intervened upon. Eq.\eqref{ic} follows as Rule 2 for action/observation exchange is satisfied. In other words, given that $(\X \indep \Y_c) | \Y_{-c},\U_x)_{G_{\underline{\y_c}}}$ is $d$-separated in $G_{\underline{\y_c}}$ in Fig.\ref{fig:g_yi_} allows the exchange from $p(\x | \hy_c, \y_{-c}, \u_x)$ to $p(\x | \y_c, \y_{-c}, \u_x)$ and since $\y = [\y_c, \y_{-c}]$ by our definition, completes the proof.
 Extensions to cases where $n>2$ as in the generative model in Sec.2.1 follows trivially through the same derivation.
 
%Splitting the two terms in Eq.\eqref{chain_rule} and applying the rules of the  
% $do$-calculus to test for $d$-Separations on each yields Eqns. \eqref{rule3}-\eqref{rule2_2}.
%%
%\begin{equation} \label{rule3}
%p( \y_j | \hy_i) = p(\y_j)
%\end{equation}
%%
%In Eq.\eqref{rule3}, Rule 3 is applied to delete the action $\hat{\y_i}$. Verification that $\y_j$ is $d$-Separated from $\hy_i$ by inspection of the subgraph shown in Fig. \ref{fig:g__yi} corresponding to $G_{\overline{\y_i}}$ is done by checking that all the paths connecting variables $\y_j$ to $\y_i$ are blocked; which is the case from the presence of the colliders at $\x,\z, \ty_j$.
%%
%\begin{equation} \label{rule2}
%p( \ty_i | \hy_i,  \y_j) = p( \ty_i | \y_i,  \y_j)
%\end{equation}
%%
%The first term in Eq.\eqref{chain_rule} can be identified instead using Rule 2 action/observation exchange in $G_{\underline{\y_i}}$. Here, by inspection of the paths in Fig.\ref{fig:g_yi_}, we see that information flow is blocked from $\y_i$ to $\ty_i$ by conditioning on $\y_j$. We thus observe that $\y_i$ is $d$-Separated from $\ty_i$ given $\y_j$. In other words, that $(\ty_i \indep \y_i | \y_j)_{G_{\underline{\y_i}} }$ is satisfied.
%
Abusing causal notation, we derive the identifiability conditions of the query $p(\z|\x, \hy_c)$ in Eq.(5) noting that all terms involved respect the causal direction.
\begin{eqnarray} 
\nonumber p(\z|\x, \hy_c) \\
= p(\x|\z, \hy_c ) p(\z|\hy_c) p(\hy_c) / p(\x,\hy_c) \label{chain_rule}\\
= p(\x|\z)   \expec_{\y_{-c},\u_x} \left [ p(\z|\y, \u_x)\right] / p(\x,\y_c) \label{der}
\end{eqnarray}
Eq.\eqref{chain_rule} follows by application of the chain rule of probability. Deletion of actions from $p(\x|\z, \hy_c)$ follows by applying Rule 3, satisfied when $d$-separation $(\X \indep \Y_c ) | \Z)_{G_{\overline{\y_c}}}$ is satisfied. By inspection of Fig.\ref{fig:g__yi} we see this is indeed the case. Also, the action $p(\hy_c)$ is by definition one and substituting the result of the conditional latent distribution $p(\z|\hy_c)$ in Eq.\eqref{ic} completes writing an equivalent expression involving only ordinary probabilities of observations for the numerator. An expression for the denominator $p(\x,\hy_c)$ follows by adding $\z$ through the law of total probability and using the chain rule as $\sum_{z}p(\x,\hy_c|\z) p(\z)$. The action/observation exchange $\sum_{z}p(x,\y_c|\z) p(\z)$ then follows by checking if $(\X \indep \Y_c) | \Z)_{G_{\underline{\y_c}}}$ is satisfied; which is indeed the case by inspection of Fig.\ref{fig:g_yi_}, completing the proof.

% In the case that of unmeasured factors.
When at least one of the generating factors, such as $\u_x$ (e.g., sensor noise), remains unmeasured, it will leave several paths (e.g., $\y_c \leftrightarrow \u_x$, $\y_c \leftrightarrow \u_x \leftrightarrow \y_{-c}$) in the collider unblocked. This means, the causal effect of $\y_c$ on $\z$ cannot be identified uniquely, but only a relaxed relationship where $p(\z|\tilde{y_c})$ may carry information correlations with both $\u_x$ and $\y_{-c}$. The strength of such correlations depends in this case on the energies of $\u_x$ relative to $\z$. But, overall the strength of these correlations has a direct impact on the severity of bias in DL models. This problem can be aggravated exponentially when the number of unmeasured generating factors increases. One of the main arguments in this research is that collider bias is prevalent in the majority of DL models designed to disentangle generative factors. These models often fall short in recognizing and effectively addressing this issue. We propose leveraging the power of causal models, specifically DAGs, to effectively incorporate a transparent and explicit model of the generative process. This integration aims to identify and mitigate the influence of colliders on disentanglement tasks. By leveraging DAGs, we can enhance the understanding and management of collider effects, improving the overall performance of disentanglement DL models.

\section{VAE+ReI Reformulation: Alignment with Causal Collider Structure}
\label{appendix:elbo} 
% Have other VAE relatedworks tackled this problem?
%Related works based on variational inference fundamentals like the VAE \cite{kingma2013auto} and even the supervised conditional VAE \cite{sohn2015learning} are susceptible to collider problem when conditioning on $\z$ as required by the likelihood function as no adjustment for the collider is provided. We recognize this as an important issue here and provide possible avenues for its resolution via the ReI framework while also providing connections to extend some of the analysis in identifiable based learning methods \cite{higgins2016beta, locatello2020weakly, trauble2021disentangled} to integrate ReI in a modularized fashion.

Through ReI, we align the VAE framework, with the causal DAG of Fig.\ref{fig:g} through a reformulation of the ELBO that accounts for the presence of a collider. The reformulation describes the learning problem in terms not of the ordinary posterior $q(\z|\x,\y)$ but rather in terms of an interventional posterior $q(\z|\x,\hy)$.
Derivation of this reformulated ELBO in Eq.(6) follows the same steps as in \cite{kingma2013auto} starting with the Kullback-Leibler (KL) divergence.

In the case of the VAE, the likelihood term is given by Eq.\eqref{likelihood} as:
\begin{equation} \label{likelihood}
	\L_{\ell}(\th, \ph; \x^{(i)}) = \expec_{q(\z|\x^{(i)})} \left [  \log p(\x^{(i)}  | \z)   \right ] 
\end{equation}
and the regularizer $\L_{\Reg}(\th, \ph; \x^{(i)})$ given in case of the standard VAE by the Kullback-Leibler (KL) divergence
\begin{equation} \label{KL}
	\L_{\Reg}(\th, \ph; \x^{(i)}) = D_{KL}( q(\z|\x^{(i)}) || p(\z))
\end{equation}
imposing a prior $p(\z)$, typically a standard Gaussian, on the approximate posterior. The $\th,\ph$ are the parameters of the encoder and decoder models, respectively, and optimized over the training dataset. A scalar $\lambda$ is typically introduced as a multiplier in front of the r.h.s. of Eq.\eqref{KL} as the regularizer strength balancing tradeoffs between the likelihood and priors. This is a parameter utilized by the $\beta$-VAE to promote the prior structure. %Other extensions without identification guarantees,  structuring the latent space by class under supervision is the conditional VAE of \cite{sohn2015learning} approximating instead, the conditional posterior $p(\z|\x,\y)$ given data pairs $\{ \x^{(i)}, \y^{(i)}\}_{i=1}^N$. 
The regularizer in Eq.\eqref{KL} is reformulated by ReI to impose disentanglement constraints using the collider model structure shown in Fig.\ref{fig:sensor_DAG}. The steps of the full derivation are:
\begin{eqnarray} 
 D_{KL} (q( \z|\x, \hy_c) || p(\z|\x, \hy_c))
= -\sum q( \z|\x, \y_c) \log p( \z|\x, \hy_c) / q( \z|\x, \y_c) \label{KL_definition}\\
= - \expec_{q(\z|\x, \y_c)} \left \{\log p(\x|\z)\right \} - \sum q( \z|\x, \y_c) \log \expec_{p(\y_{-c})} \left [ p(\z|\y)\right] / q( \z|\x, \y_c)
 + \log p(\x, \y_c). \label{expansion} 
\end{eqnarray}
Eq.\eqref{KL_definition} follows by definition of the KL divergence, while Eq.\eqref{expansion} substitutes the result from the identification adjustments for the causal query $p(\z|\x, \hy_c)$ in Eq.\eqref{der}. Based on Eq\eqref{expansion}, the ELBO can be written as in Eq.\eqref{elbo_proof}, completing its derivation. Note that in Eq.\eqref{elbo_proof} we have ommited the presence of a factor $\u_x$ and focused only on the observed factors $\y_c$.
\begin{eqnarray} 
	\nonumber \log p(\x^{(i)}, \y_c^{(i)}) \geq   \L( \th, \ph; \x^{(i)}, \y_c^{(i)}) \\
	= \expec_{q(\z|\x^{(i)}, \y_c^{(i)})} \left \{\log p(\x^{(i)}|\z)\right \}
- D_{KL} \Bigl (q( \z|\x^{(i)}, \y_c^{(i)}) || \expec_{\y_{-c}} \left [ p(\z|\y)\right] \Bigr ). \label{elbo_proof}
\end{eqnarray}

\section{Benchmark experiments}
\label{appendix:experiments}

\subsection{Generating correlations in benchmark dataset}
Correlations in the generated data where produced by the method described in \cite{trauble2021disentangled, roth2022disentanglement} with $\sigma$ quantifying the amount of correlation between factors. The smaller the $\sigma$, the stronger the correlation is, and vice versa. All pair-wise correlations where generated with $\sigma=0.1$, while a $\sigma=0.2$ was used to generate the factor correlated with all others (i.e., 1-to-all), in consistency with \cite{trauble2021disentangled, roth2022disentanglement}.

\subsection{DL model settings}
The VAE architectures used throughout the benchmarking experiments follows the implementations of \cite{locatello2020weakly, roth2022disentanglement}. The encoder consists of 2x [Conv(32,4,4) + ReLU],  2x [Conv(64,4,4) + ReLU], MLP(256), MLP(2x10). The Decoder uses: MLP(256), 2 x [upConv(64,4,4) + ReLU], 2 x [upConv(32,4,4) + ReLU], [upConv(3,4,4) + ReLU]. Inputs are images with 3 channels grouped into batches of 64 images. Training is performed using the Adam optimizer with a learning rate of 10e-4 for 300,000 training steps. In the case of Factor-VAE, the architecture includes six layers of [MLP(1000), leakyReLU] followed by an MLP(2).

In terms of the functional encoder/decoder approximators, deep model capacity is assumed to satisfy the data processing inequality with equality constraints. In other words, the mutual information $I$ between $\Y_c$ and $\Z$ is preserved relative to $\Y_c$ and $\X$ (i.e. $I(\Y_c, \Z) = I(\Y_c, \X)$). This assumption has been used in other works \cite{locatello2020weakly, mao2022causal} and justified in the VAE's objective to faithfully approximate the marginal data distribution.

\subsection{DCI}
The DCI disentanglement metric \cite{eastwood2018framework} is a measure of how each variable (or dimension) captures at most one generative factor. It can be computed for each variable or dimension $i$ as $D_i = (1-H_K(P_i))$. Here, $H_K(P_i)$ is entropy given as $H_K(P_i))=-\sum_{k=0}^{K-1} P_{ik}\log_K P_{ik}$ and $P_{ij} = R_{ij} / \sum_{k=0}^{K-1} R_{ik}$ is the probability of a learned latent variable being important for predicting a known generating factor. This later (i.e.,$R_{ij}$ ) can be computed from the  classification prediction error.

\section{Experiments with chemcam Real-World Dataset}
\label{appendix:realworld}

\subsection{Dataset details} \label{Ssec:data_details}
The ChemCam LIBS instrument \cite{wiens2012chemcam} datasets contain raw and denoised spectra obtained from a variety of targets (e.g., rocks, soil) and from reference calibration standards of known and certified chemical composition. The specific datasets we employ consists of spectrally resolved LIBS signal measurements collected on Earth in a laboratory setting from a set of $\sim$ 585 reference calibration standards \cite{Clegg:2017} and on Mars from a set of 10 reference standards of known true composition. Each target is repeatedly shot (e.g.,  50 times) following each time measurements of the full 240-905 nm LIBS signal. %This process is repeated at multiple locations for each target which amounts to $\sim$41,000 uncleaned LIBS signals. 
After collection, wavelengths within the bands [240.811,246.635], [338.457,340.797], [382.13,387.859], [473.184,492.427], [849,905.574] were ignored out consistent with practices of the ChemCam team \cite{Clegg:2017}. 

% Training and implementation details
\subsection{Training and Implementation details} \label{Ssec:implementation_details}
Hyperparameters of the DL model were set to an initial lr of 1.0, decayed after 75 epochs with cosine annealing \cite{loshchilov2017sgdr} and with \#epochs 300. Batches were constructed at each epoch from a set of $64$ shot-averaged examples randomized over the whole training set without replacement. The shot-averages where computed by averaging the LIBS signal representations over an individual target and laser shot location. This averaging is consistent with common practices of the ChemCam team \cite{Wiens:2013, Clegg:2017}. From a practical standpoint, regularization by ReI in Eq.\eqref{ReI} requires computing expectation over distributions of the generative factors. This is computationally intractable and we resorted to approximations by sampling with a limited number of samples (throughout the experiments with spectral data we used a 1000 samples) per causal relationship. This approximation resulted in some information leaks from other generating variables. This phenomenon can be observed qualitatively for example in the small peaks present in Fig.\ref{fig:rep_blocked_noisy} (from 200-500nm wavelengths).

\subsection{Additional Comparisons against DL architectures and depths}
\begin{figure}[!h]
	%\centering
	\begin{minipage}{.5\linewidth}
		\captionof{table}{Transfer Performance Comparison}
		\label{tab:comparison}
		\centering
	\begin{tabular}{lr}
	\toprule
	Architecture     & RMSE  (\% oxide)  \\
	\midrule
	FC(10)& 5.19       \\
	MLP(10)  & 5.06      \\
	MLPMixer(8) & 5.23\\
	ResNet(18) + FC(10) & 6.23 \\
	U-Net +FC(5) & 5.12 \\
	Transformer + FC(10) & 6.12 \\
	VAE + FC(10) & 4.51 \\
	%DDPM + FC(10) & xxx \\
	$\beta$-VAE +FC(10) & 4.13 \\
	F-VAE +FC(10) & 4.01 \\
	DIP-VAE +FC(10) & 4.67 \\
	ReI-VAE+FC(1) & \textbf{2.45}    \\
	\bottomrule
\end{tabular}
	\end{minipage}
\hfill
	\begin{minipage}{.5\linewidth}
	\centering
	\includegraphics[width=1.0\linewidth]{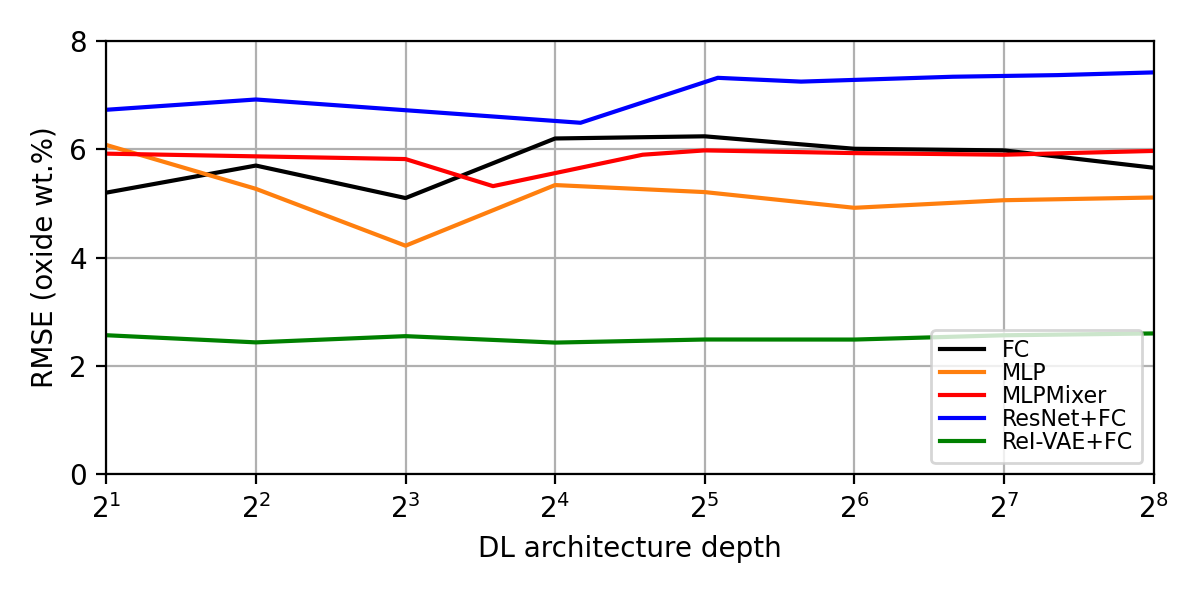}%multiview_side_2}}
	%\rule{2.8cm}{2cm}
	\captionof{figure}{Transfer performance versus DL model depth.}
	\label{fig:numlayers_comparison}
\end{minipage}%
\end{figure}

Here, we include the results of a few additional experiments in the ChemCam dataset that compare performance on out-of-distribution examples.
Table \ref{tab:comparison} provides additional results comparing performance on Earth-to-Mars transfer on a variety of DL architectures and averaged over all elements $\y \in \Rspace^n$ with $n=11$.

Comparisons include fully connected (FC), multilayer perceptron (MLP), MLP Mixer \cite{tolstikhin2021mlp}, ResNet \cite{7780459}, U-Net \cite{ronneberger2015u}, Transformers \cite{dosovitskiy2020image}, VAE \cite{kingma2013auto}, $\beta$-VAE \cite{higgins2016beta}, Factor-VAE \cite{kim2018disentangling}, DIP-VAE \cite{kumar2018variational}. Note that some of the architectures do not produce a latent representation explicitly, these are however rather trained end-to-end for prediction. The number in parenthesis next to each architecture name (e.g., FC(10)) expresses the corresponding depth of layers. The results of Table \ref{tab:comparison} show that VAE+ReI outperforms standard architectures in cases of OOD examples regardless of the inductive biases implied by the compared architectural designs. The unsupervised representation learning methods Beta-VAE, factorized VAE and DIP-VAE trained with a supervised prediction loss performed better at transfer than the standard deep learning architectures compared. However, VAE+ReI imposing disentanglement constraints via causal identification from the explicit DAG collider model, was able to outperform them all.
Fig.\ref{fig:numlayers_comparison} also shows the transfer performance as a function of DL model depth. In this case, the FC, MLP, MLPMixer and ResNet+FC networks were compared. This plot shows that VAE+ReI is capable of outperforming standard DL models which did not exhibit generalization capabilities to OOD cases regardless of depth in this case.
As a remark we would like to highlight that gains in task performance may not necessarily translate into more generalizable DL models. As evidenced by experiments, these may sometimes trick one's belief of a better model. In our case, these issues were settled through experiments evaluating the alignment of the resulting learned representations with domain knowledge. 
Finally, with regards to limitations, ReI requires a full reformulation of the learning problem when the data generation process is different from that of Fig.\ref{fig:sensor_DAG}. This human exercise of modeling the generation process through DAGs and deriving the conditions for identification of the causal effects can be time consuming. Discovering models of the generation process automatically \cite{glymour2019review} is an active area of research but this is outside the scope of this work. In some cases, causal identification for a given DAG can be more challenging to obtain or does not exist due to the presence of unobserved variables. Measurable proxies can be exploited as in \cite{kuroki2014measurement} in some of these cases, but in some others where this is not possible one has to resort to parametrization approximations which may result in entanglements of residuals between the true and sampled parameterized distributions; this, of course without identification guarantees. In the example application of chemical composition from LIBS we discussed this issue in the case of the sensor noise factor, with \cite{Wiens:2013, castorena2021deep} and without control as shown in Figs.\ref{fig:reg_blocked} and \ref{fig:rep_blocked_noisy}, respectively..

%\subsection{Concluding Remarks}
We conclude this subsection by highlighting an additional significant drawback of the state of the art methods for disentanglement in comparison to ours: they do not produce representations that align with domain knowledge. 
This limitation carries significant drawbacks specially in high-risk applications. %, including domains such as healthcare, financial risk assessment, and energy reliability. 
It also extends to other fields where the need for highly interpretable models is paramount, such as scientific research. In these contexts, the ability to understand and interpret the underlying factors driving model predictions is crucial for making informed decisions and ensuring the reliability and safety of the outcomes. Addressing this limitation becomes particularly vital in such applications. %These domains often demand cautious adoption of DL models due to the potential consequences of biased predictions. As a result, the integration of DL approaches has progressed at a slower pace in these areas, emphasizing the need for through consideration of the limitations and potential risks associated with these models.

							% Appendix

%%%%%%%%%%%%%%%%%%%%%%%%%%%%%%%%%%%%%%%%%%%%%%%%%%%%%%%%%%%%

\end{document}